\documentclass[sigconf, screen, balance=false]{acmart}
\usepackage{algorithm}
\usepackage{algorithmic}
\usepackage[title]{appendix}
\usepackage{xcolor}
\usepackage{booktabs}
\usepackage{multicol}
\usepackage{mathtools}
\usepackage{multirow}
\usepackage{adjustbox}
\usepackage{tcolorbox}
\tcbuselibrary{listings,skins,breakable}
\usepackage{listings}
\usepackage{fontenc}
\usepackage{fontawesome5}
\usepackage{pifont}
\usepackage{url}
\usepackage{array}
\newtheorem{proposition}{Proposition}
\definecolor{up_color}{RGB}{0, 108, 0}    %
\definecolor{down_color}{RGB}{230, 15, 15}  %
\definecolor{darkgreen}{rgb}{0, 0.492, 0}

\newcommand{\calO}{\mathcal{O}} %
\newcommand{\calS}{\mathcal{S}} %
\newcommand{\calI}{\mathcal{I}} %

\definecolor{promptbg}{RGB}{248, 249, 250}
\definecolor{promptframe}{RGB}{52, 73, 94}
\definecolor{prompttitle}{RGB}{41, 128, 185}

\definecolor{prompttitle-grey}{RGB}{200, 200, 200}
\definecolor{prompttitle-green}{RGB}{39, 174, 96}
\definecolor{prompttitle-red}{RGB}{192, 57, 43}
\definecolor{prompttitle-purple}{RGB}{142, 68, 173}
\definecolor{prompttitle-orange}{RGB}{230, 126, 34}
\definecolor{prompttitle-teal}{RGB}{22, 160, 133}

\newcommand{\arrowraise}{0.08ex}

\newcommand{\UpG}{\textcolor{green!60!black}{\raisebox{\arrowraise}{\faAngleUp}}}

\newcommand{\UpUpG}{\textcolor{green!60!black}{\raisebox{\arrowraise}{\faAngleDoubleUp}}}

\makeatletter
\newcommand{\appendixcovertitle}{%
    \twocolumn[{%
        \begin{center}
            {\LARGE\bfseries Appendices for \@title\par}
            \vspace{1em}
        \end{center}
    }]%
}
\makeatother

\newtcolorbox{systempromptbox}[3][]{
    enhanced,
    breakable,
    colback=promptbg,
    colframe=promptframe,
    coltitle=white,
    fonttitle=\bfseries\sffamily,
    title={#2},
    attach boxed title to top left={yshift=-2mm, xshift=5mm},
    boxed title style={
        colback=#3,
        rounded corners,
    },
    top=4mm,
    left=3mm,
    right=3mm,
    bottom=3mm,
    arc=2mm,
    boxrule=0.5pt,
    shadow={1mm}{-1mm}{0mm}{black!20},
    #1
}

\newtcolorbox{systempromptboxappendix}[3][]{
    enhanced,
    breakable,
    colback=promptbg,
    colframe=promptframe,
    coltitle=white,
    fonttitle=\bfseries\sffamily,
    title={#2},
    attach boxed title to top left={yshift=-2mm, xshift=5mm},
    boxed title style={
        colback=#3,
        rounded corners,
    },
    top=3mm,
    left=2mm,
    right=2mm,
    bottom=2mm,
    arc=2mm,
    boxrule=0.5pt,
    #1
}

\newtcolorbox{systempromptverbatim}[1][]{
    enhanced,
    breakable,
    colback=promptbg,
    colframe=promptframe,
    coltitle=white,
    fonttitle=\bfseries\sffamily,
    fontupper=\ttfamily\small,
    title=System Prompt,
    attach boxed title to top left={yshift=-2mm, xshift=5mm},
    boxed title style={
        colback=prompttitle,
        rounded corners,
    },
    top=4mm,
    left=3mm,
    right=3mm,
    bottom=3mm,
    arc=2mm,
    boxrule=0.5pt,
    #1
}

\newcommand{\changeindicator}[1]{%
  \ifdim #1pt >0pt%
    \textcolor{up_color}{\tiny{\, +#1\%}}%
  \else%
    \textcolor{down_color}{\tiny{\, #1\%}}%
  \fi%
}

\newcommand{\negchangeindicator}[1]{%
  \ifdim #1pt >0pt%
    \textcolor{down_color}{\tiny{\, +#1\%}}%
  \else%
    \textcolor{up_color}{\tiny{\, #1\%}}%
  \fi%
}

\definecolor{darkred}{RGB}{195,16,16}

\AtBeginDocument{%
  }

\setcopyright{acmlicensed}
\copyrightyear{2026}
\acmYear{2026}
\acmISBN{979-8-4007-2213-4/2026/11}
\acmDOI{10.1145/3767308.3835119}
\acmConference[MM '26]{Proceedings of the 34th ACM International Conference on Multimedia}{November 10--14, 2026}{Rio de Janeiro, Brazil}
\acmBooktitle{Proceedings of the 34th ACM International Conference on Multimedia (MM '26), November 10--14, 2026, Rio de Janeiro, Brazil}

\begin{document}

\title{On Semiotic-Grounded Interpretive Evaluation of Generative Art}

\author{\href{https://j-rx.com}{Ruixiang Jiang}}
\email{rui-x.jiang@connect.polyu.hk}
\affiliation{
  \institution{The Hong Kong Polytechnic University}
  \city{Hong Kong}
  \country{China}
}

\author{\href{https://chenlab.comp.polyu.edu.hk/}{Chang Wen Chen}}
\email{chen.changwen@polyu.edu.hk}
\affiliation{
  \institution{The Hong Kong Polytechnic University}
  \city{Hong Kong}
  \country{China}
}

\begin{abstract}
\emergencystretch=1em
Interpretation is essential to deciphering the language of art: audiences communicate with artists by recovering meaning from visual artifacts. However, current Generative Art (GenArt) evaluators primarily emphasize surface-level image quality or literal prompt adherence, failing to assess the deeper symbolic or contextual meaning intended by the creator. We formalize this observation using Peircean computational semiotic theory that models Human-GenArt Interaction (HGI) as cascaded semiosis. This framework reveals that artistic meaning is conveyed through iconic, symbolic, and indexical modes, whereas appearance-centered evaluators tend to emphasize the iconic mode. Building on this account, we propose SemJudge, an interpretation-centric evaluation framework that uses a Hierarchical Semiosis Graph (HSG) to reconstruct meaning conveyance from prompt to generated artifact. Experiments on an interpretation-intensive benchmark, SemiosisArt, show that SemJudge aligns more closely with human judgments than existing baselines. User studies further demonstrate that SemJudge produces deeper, more insightful artistic interpretations. We hope these findings pave the way for GenArt to move beyond the generation of ``pretty'' images toward a medium capable of understanding and expressing complex human experience. Project page: \url{https://github.com/songrise/SemJudge}
\end{abstract}

\thanks{Author manuscript with appendices. The version of record appeared in the Proceedings of the 34th ACM International Conference on Multimedia (MM~\textquotesingle26), 2026, pp. 3814--3823. \protect\url{https://doi.org/10.1145/3767308.3835119}.}
\makeatletter
\pretocmd{\@mkbibcitation}{%
  \if@firstcolumn\newpage\fi
  \noindent\begin{minipage}{\columnwidth}
  \centering
  \includegraphics[width=\linewidth]{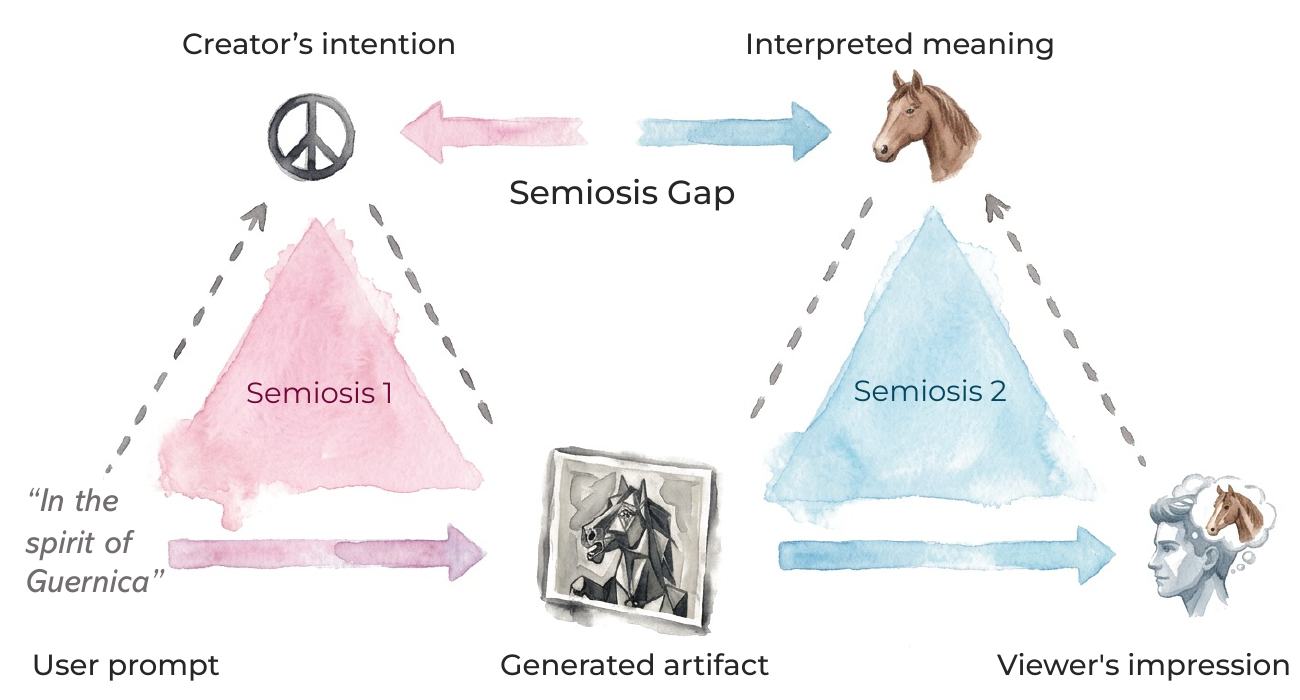}
  \captionof{figure}{\textbf{HGI as cascaded semiosis.} We model HGI as a chain of meaning-making processes. A creator articulates an intention through a prompt, which the model interprets to generate an artifact. A viewer subsequently interprets the artifact and reconstructs a meaning that may differ from the creator's original intention.}
  \label{fig:genart_mapping}
  \end{minipage}\par
}{}{\PackageError{semjudge}{Unable to place the first-page figure}{}}
\makeatother
\maketitle

\section{Introduction}
\label{sec:introduction}

\begin{quote}
\emph{``To see something as art requires something the eye cannot descry.''}

\hfill — Arthur C. Danto, \emph{``The Artworld''} \cite{danto1964artworld} 
\end{quote}

\emergencystretch=1em

At its core, art is an act of meaning-making~\cite{goodman1976languages,langer1953feeling}.  What distinguishes a painting from a photograph of the same scene is often not fidelity to appearance alone, but the deliberate articulation of artistic intent through metaphor, symbolism, abstraction, and convention~\cite{danto1964artworld,goodman1976languages,danto1981transfiguration}. Interpretation is therefore central to aesthetic engagement. Yet existing Generative Art (GenArt) evaluation methods remain heavily fixated on what ``the eye can descry'' — measuring realism~\cite{heusel2017gans,wright2022artfid}, prompt-image alignment~\cite{radford2021learning,ku2024viescore}, or generic visual appeal~\cite{kirstain2023pick,schuhmann2022laion}, while leaving the deeper artistic meaning largely untouched. Accordingly, these evaluators are often misaligned with human judgments of trained viewers~\cite{chamberlain2018putting, epstein2023art,samo2023artificial,van2025human,hullman2023artificial}. We identify two contributing factors to this mismatch: 

\textbf{Gap 1: Artistic meaning is not reducible to surface appearance.} Instead, it is often encoded through non-literal strategies such as juxtaposition, abstraction, and metaphorical cues~\cite{danto1964artworld,goodman1976languages} that diverge deliberately from surface appearance. For example, the impact of Picasso's \emph{Guernica} comes less from resembling a photorealistic scene of war and more from how tonal harshness, fragmentation, and distorted figures convey moral outrage and an anti-war stance~\cite{chipp1988picasso}. However, appearance-centric evaluators risk conflating artistic meaning with surface quality, rewarding visual fidelity or aesthetic allure as proxies for artistic quality.

\textbf{Gap 2: Artistic intent is not reducible to literal prompt wording.}
Just as meaning may be communicated indirectly through an artwork, a creator's intention may be expressed indirectly through language~\cite{grice1975logic}. GenArt prompts do not always function as exhaustive descriptions of the desired image; they may instead operate as artistic directions that specify a theme, atmosphere, motif, or intended effect~\cite{vodrahalli2023artwhisperer,chang2023prompt}.
A prompt such as ``\emph{in the spirit of Guernica},'' for instance, does not prescribe a visual layout. It invokes a historically and artistically situated mode of expression that the generator must interpret.
A capable GenArt system should therefore be able to interpret such indirect prompts and render these intended effects through strategies such as exaggeration or abstraction. Existing evaluators, however, often assess direct text--image alignment and may therefore underrepresent this interpretive step.

What is missing, then, is not simply stronger visual perception, but an explicit account of the \textbf{interpretive process} connecting intention, prompt, generated artifact, and judgment. In particular, when artistic meaning is conveyed through metaphor, symbolism, or convention rather than literal resemblance, evaluation can no longer rely on appearance alone~\cite{goodman1976languages,cassirer1953philosophy,langer2009philosophy,danto1981transfiguration}. To model this relation, we draw on \textbf{semiotics}, which has long informed both art theory~\cite{morgan1955icon,bal1991semiotics,langer1953feeling} and human--computer interaction (HCI)~\cite{de2005semiotic}, and provides a vocabulary for studying how meaning is communicated across representations.
In this paper, we use semiotics to model Human--GenArt Interaction (HGI) as a process in which a creator's intention is articulated through prompts and realized in generated artifacts (e.g. images). This perspective also helps characterize the iconicity bias of conventional metrics, which can cause them to diverge from human judgments of non-representational artworks.

{\emergencystretch=1em
Building on this theory, we propose \textbf{SemJudge}, an interpretation-centric HGI evaluator that reconstructs how meaning is conveyed from the prompt to the generated artifact, rather than merely scoring surface-level alignment or visual appeal. SemJudge uses \textbf{Hierarchical Semiosis Graphs (HSGs)} to represent linked meaning units and connect interpretive claims to prompt spans and image regions. Experiments on our proposed \textbf{SemiosisArt} dataset show that SemJudge aligns more closely with human judgments and yields more informative, auditable, and in-depth interpretations of artistic meaning embedded in generated artifacts. We summarize our contributions as follows:
}
\begin{enumerate}
    \item \textbf{A semiotic framework for HGI.} We formalize HGI as cascaded semiosis and characterize why appearance-centered metrics may become less reliable when artistic meaning is conveyed indirectly.
    \item \textbf{An interpretation-centric evaluator.} We introduce \textbf{SemJudge}, which uses \textbf{Hierarchical Semiosis Graphs} to reconstruct meaning conveyance and ground interpretive claims in prompt spans and image regions.
    \item \textbf{A benchmark and empirical analysis.} We show that SemJudge aligns more closely with human judgments and yields more informative, edifying interpretations than strong baselines.
\end{enumerate}

\section{Related Work} 
\subsection{GenArt Evaluation} \emergencystretch=1em Early evaluation metrics primarily emphasized realism, often quantified through the distance (or divergence) between generated and real-image distributions. Metrics such as Inception Score~\cite{salimans2016improved}, FID~\cite{heusel2017gans} and ArtFID~\cite{wright2022artfid} fall into this category. With the rise of text-conditional generation, GenArt systems increasingly incorporated text-image alignment~\cite{radford2021learning}. Subsequent work enhanced alignment scoring through human-preference tuning, giving rise to models such as PickScore~\cite{kirstain2023pick} and HPS~\cite{wu2023human}. Although these preference models capture aspects of generic visual appeal, they typically provide an aggregate score with limited insight into the basis for that judgment. More recently, Question Generation and Answering (QG/A) models have enabled more interpretable and structured evaluation~\cite{hu2023tifa,cho2024davidsonian,ku2024viescore}. While existing GenArt evaluators are well suited to measuring appearance-level realism and visual attractiveness, deeper artistic meaning has received comparatively less attention~\cite{hullman2023artificial,jiang2025multimodal}. This paper adopts a semiotics-theoretical lens to (1) characterize the limitations of appearance-driven metrics and (2) inform the design of a meaning-driven evaluation framework that supports explicit interpretation.

\subsection{GenArt Interpretation and Theory of Art}  \emergencystretch=1em
Art interpretation has long been grounded in art theory, notably Panofsky's iconological reading of visual motifs in religious art~\cite{Panofsky1955}. Semiotic traditions further treat art more broadly as a system of signs~\cite{langer1953feeling,goodman1976languages,shapiro1974intention}. Existing computational methods mainly approach this task through retrieval~\cite{garcia2018read,bleidt2024artquest,wang2025artrag} or model adaptation to curated datasets~\cite{bin2024gallerygpt,huang2024aesexpert,cao2025artimuse}. While these approaches have shown promise for interpreting historical paintings, their applicability to GenArt evaluation may be limited in two respects. First, they are typically evaluated on canonical artworks that are widely represented in pretraining corpora, making it difficult to disentangle interpretive competence from possible memorization~\cite{rudman2025forgotten,zheng2024thinking,jiang2025multimodal}. Second, their focus on the visual artifact places comparatively less emphasis on how meaning is conditioned by prompt intent and realized through the HGI process. Recently, ArtCoT~\cite{jiang2025multimodal} demonstrated the effectiveness of zero-shot MLLMs for aesthetic judgment, though it explicitly treats symbolic art interpretation as a form of hallucination to be suppressed. In the context of interpreting GenArt, this work pioneers the use of semiotics to interpretively decipher meaning-making in the entire human--GenArt co-creation process.

\subsection{Computational Semiotics} Computational semiotics studies how semiotic concepts can be formalized and computed for meaning-driven intelligence systems. Work in this area has long informed HCI, where semiotic models help explain how users interpret interfaces and how meaning is negotiated in interaction~\cite{de2005semiotic,de2009semiotic,morra2024semiotic}.
More recently, semiotic perspectives have been used to examine the behavior and limitations of contemporary AI systems, suggesting that current models may rely heavily on surface-level patterns without explicitly modeling sign relations or processes of meaning-making~\cite{picca2025not,valdez2024semiotics,morra2024semiotic,kuang2025express}. These observations motivate the need for evaluation frameworks that attend more directly to meaning conveyance. In this paper, we apply Peircean semiotic theory to GenArt evaluation and propose an interpretive evaluation framework that examines how meaning is conveyed from the prompt to the generated artifact.

\section{Human-GenArt Interaction as Semiosis}
\label{sec:theory}

\begin{figure*}[!t]
    \centering
    \includegraphics[width=\linewidth]{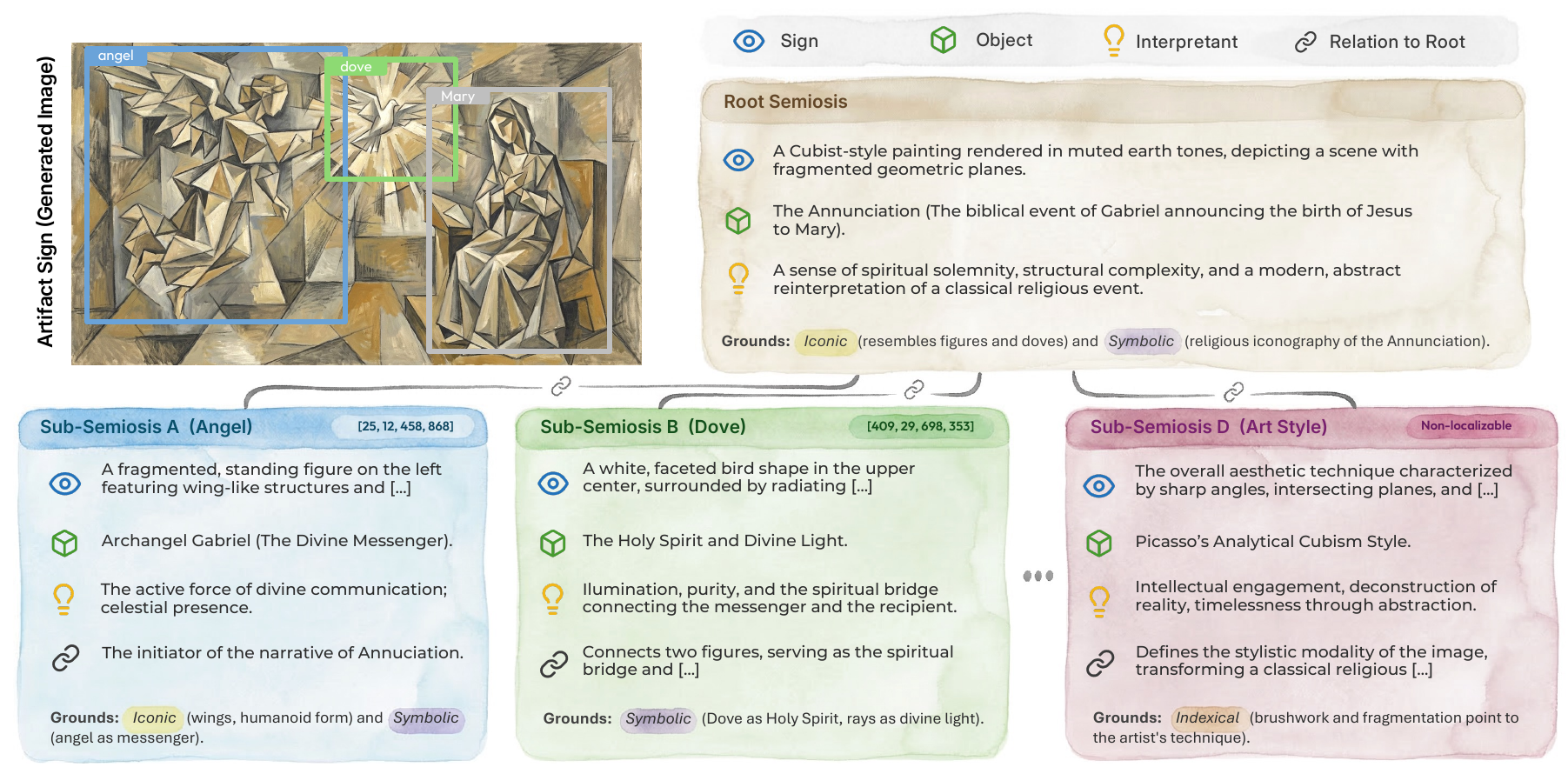}
    \caption{\textbf{HSG of generated artifact.} We show the image with bounding boxes (top-left), its global semiosis (top-right), and sub-semioses (bottom), constructed by an MLLM in zero-shot. The HSG provides a structured interpretation of the Annunciation motif in the abstract painting. Best viewed in color. }
    \label{fig:main}
\end{figure*}

This section introduces Peircean semiotic concepts and formal abstractions through which we model HGI.

\subsection{Formulating Peircean Triadic Semiosis}

\label{ssec:triad}

\textbf{Semiosis and its Components.} Peircean semiotics characterizes \emph{semiosis} (i.e., the process of signification) as an irreducibly triadic relation among a sign (or \emph{Representamen} in Peirce's terminology) $s\in\mathcal S$, an object $o\in\mathcal O$, and an interpretant $i\in\mathcal I$~\cite{peirce1991peirce}. We denote this basic unit as an \emph{atomic semiosis}
\begin{equation}
\xi := (o,s,i)\in\calO\times\calS\times\calI.
\end{equation}
Here, the sign is the perceptible form being interpreted (e.g., a prompt or an image), the interpretant is the meaning constructed by an interpreter, and the object is the underlying referent or intended content (e.g., motif).

\textbf{Interpreted-as Relationship $s\to i$.}
In Peircean semiotics, interpretation depends on the interpreter. To make this explicit, we model an interpreter $\eta\in\mathcal{H}$, where $\mathcal{H}$ denotes the space of possible interpreters (e.g., humans or computational models), and write the interpreted-as relation as 
$i=\eta(s)$.

\textbf{Grounds and Modes of Sign--Object Relation.}
A sign stands for an object on the basis of a \emph{ground}: the features or relations in virtue of which an interpreter takes the sign to represent that object. We formalize a ground as $g\subseteq\Gamma$, where $\Gamma$ denotes the universe of possible grounds. Depending on the nature of this grounding relation, a ground may operate \emph{iconically} through shared qualities or resemblance, \emph{symbolically} through a rule, habit, or convention, or \emph{indexically} through a factual or existential connection~\cite{peirce1992essential}. These modes are not mutually exclusive; a single sign may be grounded in several modes simultaneously and to different degrees. This is especially common in art, where meaning may emerge jointly through perceptual resemblance, culturally established symbolism, and contextual or historical connections~\cite{goodman1976languages,elkins1999domain}.

\textbf{Stands-for Relationship $s \to o$}. Peirce further distinguishes between the \emph{dynamic object} ($o$), the canonical intent or reality driving the sign (e.g., the creator's latent goal), and the \emph{immediate object} ($\hat{o}$), the object as specifically represented within the sign~\cite{peirce1991peirce}. Computationally, we treat the semiotic ground as a computational evidence layer $g = E(s)$. We then define the stands-for relationship as $\hat{o}=\sigma(g;\eta)$, where an interpreter reconstructs the immediate object from the evidence layer.

\textbf{Cascaded Semiosis.}
Meaning-making in HGI is iterative: an interpretant produced at one stage can be reified as the next sign and interpreted again. We call this process \emph{cascaded semiosis}. Formally, an $N$-round cascade is
\begin{equation}
\begin{aligned}
\mathcal{C}^{(N)}
&:= \big[(\xi^{(1)},\eta^{(1)}) \rightarrow (\xi^{(2)},\eta^{(2)}) \rightarrow \cdots \rightarrow (\xi^{(N)},\eta^{(N)})\big], \\
&\text{s.t.}\quad s^{(n+1)}=\rho^{(n)}(i^{(n)}) \quad \forall\, n\in\{1,\ldots,N-1\},
\end{aligned}
\label{eq:cascaded}
\end{equation}
where $\xi^{(n)}=(o^{(n)},s^{(n)},i^{(n)})$ denotes the $n$-th atomic semiosis, $\eta^{(n)}$ denotes the interpreter, and $\rho^{(n)}$ denotes the reification process (e.g., image generation) from interpretation to the subsequent sign.

\subsection{Human-GenArt Interaction as Semiosis}
\label{ssec:cascade}
Semiotics has long provided art theory with a principled vocabulary for meaning conveyance~\cite{morgan1955icon,cassirer1953philosophy,peirce1991peirce,langer1953feeling,shapiro1974intention}. We model human--GenArt interaction as a cascaded semiosis. To understand this, consider a typical single-step generation workflow. A human user comes to the generator with an intended meaning or goal $o^{(1)}$, which is not directly observable to the GenArt system. To act on this intent, the user writes a prompt $s^{(1)}$, which can be textual or multimodal. The generator then functions as an interpreter $\eta^{(1)}$, which maps $s^{(1)}$ into its internal representation $i^{(1)}$ (e.g., text encoding). Based on this interpretation, the model synthesizes an artifact $s^{(2)} = \rho^{(1)}(i^{(1)})$. The resulting artifact is then interpreted—and, in an evaluation setting, assessed—by a subsequent interpreter $\eta^{(2)}$, such as a human user or automated evaluator. Thus, even the simplest HGI setting can be modeled as a two-stage cascaded semiosis $\mathcal C^{(2)}$, as visualized in Figure~\ref{fig:genart_mapping}. Repeated rounds of generation and interpretation extend this notation to longer cascades.

\section{Semiotics-Grounded GenArt Evaluation} 
\label{sec:computational_semiotics} 
This section formalizes a theoretical limitation of existing GenArt evaluation systems. We then introduce SemJudge, a semiotics-grounded interpretive evaluator.

\subsection{Semiosis Quality Measure} Under a semiotic view, evaluating the quality of HGI amounts to assessing the quality of the semiosis induced by human--GenArt interaction. Accordingly, we define the theoretical quality of an $N$-round semiosis $\mathcal{C}^{(N)}$ as the distance between its initial and final dynamic objects:

\begin{equation}
 Q_{\mathcal{C}^{(N)}} :=  -\Delta_o(o^{(1)},o^{(N)}),
    \label{eq: theoretical_Qc}
\end{equation}
where $\Delta_o$ is a distance metric in $\calO$, and smaller distance indicates higher quality. Because dynamic objects are latent~\cite{peirce1991peirce}, we approximate quality through interpreter-reconstructed immediate objects. The resulting \emph{empirical quality measure} of an $N$-round semiosis is defined as:

\begin{equation}
\begin{aligned}
\hat{o}^{(n)} &= \sigma(E(s^{(n)});\eta) \\
\hat{Q}_{\mathcal{C}^{(N)}}^{\eta} &:= -\Delta_o\left(\hat{o}^{(1)},\hat{o}^{(N)}\right),
\end{aligned}
\label{eq:conditioned_Qc}
\end{equation}
which yields an interpreter-conditioned and operationally estimable HGI quality in $\mathcal{C}^{(N)}$.

\subsection{A Semiotic View of Conventional GenArt Metrics}\label{sec:demystify}

\textbf{Semiotic principle.} Our framework provides an account of why GenArt evaluators can diverge systematically from human judgment~\cite{epstein2023art,kirstain2023pick,van2025human} even when literal prompt matching and visual attractiveness appear strong. We summarize this potential limitation in the following proposition:

\begin{proposition}[Interpretive Principle: iconicity mismatch degrades semiosis quality.]
\label{prop:quality_principle}
Let $\alpha(s^{(n)},\eta^{(n-1)})\in[0,1]$ denote the \emph{intended iconicity} of sign $s^{(n)}$ as encoded by its creator $\eta^{(n-1)}$, and let $\alpha(s^{(n)},\eta^{(n)})\in[0,1]$ denote the \emph{interpreted iconicity} as inferred by the subsequent interpreter $\eta^{(n)}$. We formalize the following semiotic principle:
\begin{equation}
\left|\alpha(s^{(n)},\eta^{(n-1)})-\alpha(s^{(n)},\eta^{(n)})\right| \uparrow
\quad \Longrightarrow \quad
Q_{\mathcal{C}^{(N)}} \downarrow .
\label{eq:quality_principle}
\end{equation}
That is, greater mismatch between intended and interpreted iconicity is expected to correspond to lower semiosis quality. This proposition is consistent with art-historical accounts that distinguish figurative resemblance from symbolic function~\cite{gombrich1995story}. For example, Picasso's \emph{Guernica} employs stylized forms and symbolic conventions to convey meaning~\cite{chipp1988picasso}. An audience or an evaluator that prioritizes literal resemblance may underweight these symbolic relations, increasing the divergence between the intended and interpreted objects.
\end{proposition}

Appendix~\ref{app:theory} provides a brief theoretical justification from Peircean semiosis and our formulation. We use this principle to motivate the empirical iconicity-bias analysis in Section~\ref{sec:icon_bias}.

\textbf{Framing Existing Evaluators:} Existing GenArt metrics typically operate in canonical ground space without interpretation. While this can be a reasonable proxy for quality when a sign is predominantly iconic, it becomes semiotically unreliable when meaning depends on symbolic or indexical interpretation. Depending on whether the evaluator is aware of the user input prompt, most metrics fall into two families:

\begin{enumerate}
    \item \textbf{Context-conditioned metrics (prompt-aware).} These metrics assess how well the extracted grounds of a generated artifact match the input prompt (e.g., CLIP~\cite{radford2021learning}, PickScore~\cite{kirstain2023pick}, and MLLM-based scoring~\cite{ku2024viescore}) by computing a distance between prompt and image ground representations:
    \begin{equation}
 Q(s^{(n)};s^{(1)}) \;=\; -\Delta_g\!\left(E_i(s^{(1)}),\, E_o(s^{(n)})\right),
        \label{eq:ground_cond}
    \end{equation}
 where $E_i, E_o$ are ground extractors (potentially different encoders for text and image) and $\Delta_g$ is a generic distance in the induced ground space.

    \item \textbf{Context-free metrics (prompt-agnostic).} These metrics evaluate global realism, quality, or aesthetics without reference to the prompt (e.g., FID, aesthetic predictors) by comparing the artifact to an \emph{idealized ground prior} $g^\star$ precomputed or learned from data:
    \begin{equation}
 Q(s^{(n)};\varnothing) \;=\; -\Delta_g\!\left(g^\star,\, E(s^{(n)})\right).
        \label{eq:ground_free}
    \end{equation}
\end{enumerate}

Despite their differences, both families optimize agreement in ground space rather than recovery in object space.

The shared limitation of these metrics is therefore not ground-space comparison itself, but treating it as a universal proxy for semiosis quality. As Proposition~\ref{prop:quality_principle} shows, this proxy can fail when intended iconicity diverges from interpreted iconicity, which is common in art. This can happen both in generation (e.g., the user expresses a prompt symbolically but the model interprets it iconically) and in evaluation (e.g., the evaluator has an iconicity bias and fails to recognize symbolic meaning). Both cases lead to low human satisfaction despite high 
ground-space scores, potentially explaining the observed divergence 
between GenArt and real-art evaluations; we examine this relationship empirically in Section~\ref{sec:experiment}.

\subsection{The SemJudge}
\label{sec:semjudge}

\textbf{Hierarchical Semiosis Graph.} Our original formulation in Equation~\ref{eq:cascaded} views the prompt $s^{(1)}$ and generated artifact $s^{(N)}$ holistically as atomic units in semiosis. For \emph{practical} HGI evaluation, however, it is often useful to make the internal structure of signs explicit, since both prompts and images exhibit rich compositional organization (e.g., sentence structure, entities/attributes, spatial relations, and global style)~\cite{biederman1987recognition,partee1984compositionality}.

To capture this compositional structure, we introduce the \emph{Hierarchical Semiosis Graph} (HSG), a scene-graph-inspired representation whose nodes encode atomic semioses rather than only entities and their relations. Specifically, an HSG is a directed graph $\operatorname{HSG}(s)=(\mathcal V,\mathcal E)$ where each node $v\in\mathcal V$ is an atomic semiosis $\xi$. The root semiosis $(\hat{o},s,i)$ provides a global-level analysis and is connected to interpretable \emph{sub-semioses}, which analyze the meaning of sub-signs in $s$. Edges $e\in\mathcal E$ between global and sub-semioses encode their relations (e.g., supports/elaborates, contrasts), thereby making explicit both (a) \emph{what} meanings are present locally and (b) \emph{how} they interact to form global intent. Following semiotic theory~\cite{peirce1992essential,eco1979theory}, we represent all components of the HSG in natural language. This representation also supports both human understanding and downstream MLLM-based judgment and interpretation.

We further distinguish \emph{non-localizable} sub-semioses (e.g., overall style, genre, non-figurative representations) from \emph{localizable} sub-semioses (e.g., figures, objects)~\cite{kress2020reading,gatys2016image}. In implementation, localizable sub-semioses are grounded to explicit evidence: text spans in $s^{(1)}$ and bounding boxes in $s^{(N)}$, enabling interpretable, auditable, and fine-grained analysis. Figure~\ref{fig:main} presents an example of HSG for a generated artifact.

\textbf{Operationalizing Object-Space Semiosis Quality.}
Unlike canonical ground-space metrics, SemJudge explicitly reconstructs the two-round cascaded semiosis induced by HGI. Specifically, we represent a prompt--artifact interaction as the two-stage chain:
\begin{equation}
\mathcal{C}^{(2)} \;\approx\;
\big[\operatorname{HSG}(s^{(1)}) \rightarrow \operatorname{HSG}(s^{(2)})\big],
\end{equation}
where $s^{(1)}$ is the prompt and $s^{(2)}$ is the generated artifact.

SemJudge assesses relative semiosis quality under a 2AFC protocol.
Given two artifacts $s^{(2)}_a$ and $s^{(2)}_b$ generated from the same prompt $s^{(1)}$, SemJudge outputs two reconstructed semioses, node-level evidence grounding $\mathcal{L}$, and a binary judgment $\hat{y}\in\{a,b\}$:
\begin{equation}
\operatorname{SemJudge}(s^{(1)}, s^{(2)}_a, s^{(2)}_b)
\rightarrow
\big(\mathcal{C}^{(2)}_a,\mathcal{C}^{(2)}_b,\mathcal{L},\hat{y}\big).
\end{equation}
Let $\tilde{\mathcal V} := \mathcal V(\mathcal{C}^{(2)}_a)\uplus \mathcal V(\mathcal{C}^{(2)}_b)$ be the disjoint union of HSG nodes from both semioses.
Evidence grounding is a collection of node-cited natural-language rationales:
\begin{equation}
\mathcal{L} := \{(v,\ell_v)\mid v\in \tilde{\mathcal V}\},
\end{equation}
where $\ell_v$ is an interpretable explanation with an explicit citation to the corresponding semiosis node $v$.

\section{SemiosisArt}
\begin{figure}
    \centering
    \includegraphics[width=\linewidth]{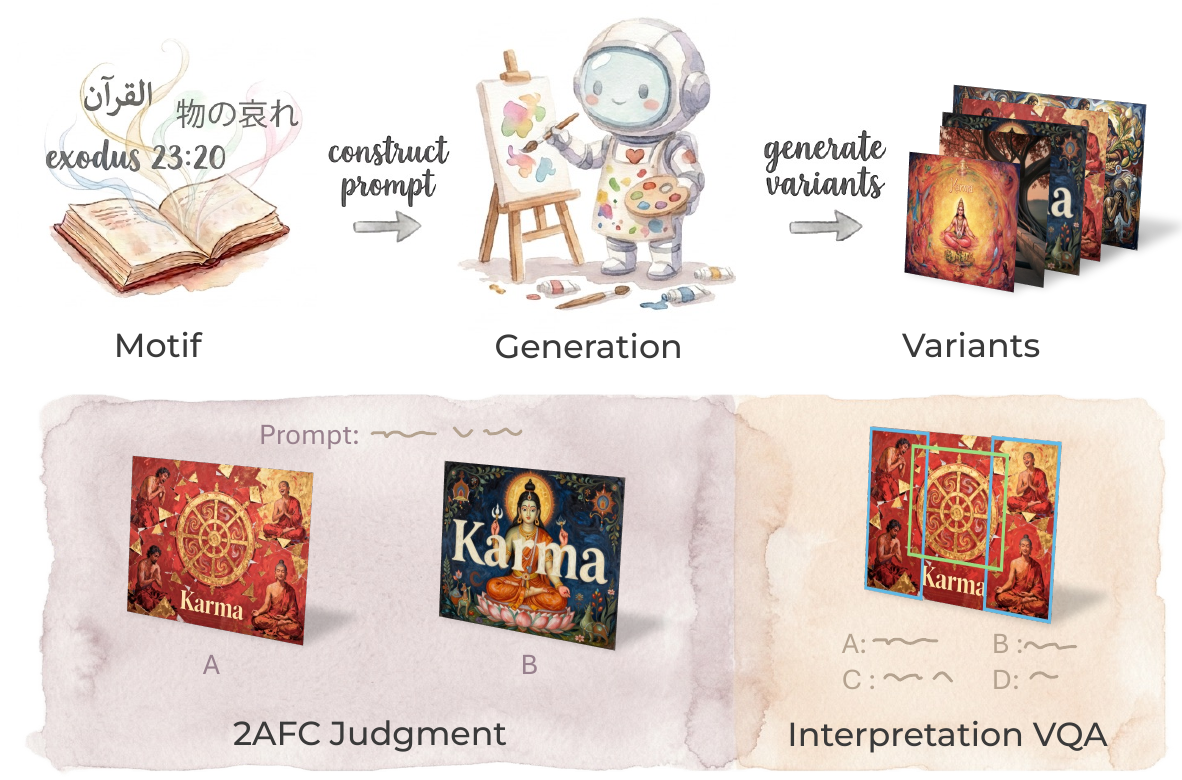}
    \caption{\textbf{SemiosisArt Construction.} Top: we construct prompts from canonical motifs and generate images using various models. Bottom: we use two task formats: 2AFC for relative judgment and VQA for fine-grained interpretation.}
    \label{fig:dataset_design}
    \vspace{-0.4cm}
\end{figure}

\paragraph{Challenge.} Existing GenArt benchmarks (e.g., AGIQA-3k~\cite{li2023agiqa}, GenAI-Bench~\cite{li2024genai}) and art-historical interpretation datasets (e.g., SemArt~\cite{garcia2018read}, VQArt-Bench~\cite{alfarano2025vqart}) are ill-equipped to evaluate artistic meaning conveyance in GenArt. First, the majority of GenArt benchmarks emphasize \emph{iconic} generation tasks. This bias toward iconic prompts and appearance-level quality makes them poorly aligned with our goal of measuring meaning-level quality in \emph{symbolic} and \emph{indexical} art forms. Art-historical datasets, on the other hand, consist of canonical artworks that are already widely covered in MLLM pretraining corpora, making strong performance difficult to disentangle from memorization rather than genuine interpretation. We therefore collect a new dataset for benchmarking HGI semiosis quality, with a focus on non-iconic generation tasks.

\paragraph{Dataset design.} Annotation subjectivity is the main challenge in constructing a dataset for interpretive semiosis quality assessment. We address this by \textbf{motif grounding} and \textbf{quality control}. First, during construction, we collaborate with $m_1=12$ experts to reduce interpretive arbitrariness by using canonically grounded interpretations. This is achieved by anchoring HGI tasks to canonical motifs with established roots in tradition (e.g., iconology, culture, theology, literature). These motifs support a degree of shared interpretive consensus because their meanings are grounded in cultural convention rather than individual preference~\cite{gemtou2010subjectivity,Panofsky1955}. Second, we use a strict quality-control process. For each 2AFC task, the majority judgment of the expert panel is taken as the reference answer. We additionally collected $38,155$ non-expert judgments and used them to filter out tasks with low inter-subjective agreement, achieving an inter-annotator agreement of 0.58 (Cohen's $\kappa$). The final dataset contains $187$ HSG instances, with $935$ images generated from $16$ generative models. Further details on the dataset construction and quality control are provided in Appendix~\ref{appendix:dataset}.

\paragraph{Tasks.} SemiosisArt provides two task formats: a 2AFC judgment task and a VQA task, as illustrated in Figure~\ref{fig:dataset_design}. The judgment task is the main format. Specifically, we use \textbf{2-Alternative Forced Choice (2AFC)}, which is generally considered more reliable than averaged Likert scales (i.e., Mean Opinion Score) for subjective judgments~\cite{mantiuk2012comparison,maydeu2010item}. We additionally consider the \textbf{Visual Question Answering} (VQA) format. The two formats are complementary: 2AFC captures relative quality judgments at the instance level, while VQA probes whether models can perform fine-grained semiotic interpretation. The 2AFC task contains 1,870 comparative judgments, while the VQA task contains 600 questions.

\section{Experiment and Analysis}
\label{sec:experiment}
\subsection{Experiment Settings}

\begin{table*}[ht]
    \centering    
    \caption{\textbf{Correlation and interpretation QA results across evaluators.} KRCC, SRCC, and CCC measure alignment with human judgments of HGI semiosis quality; VQA Acc measures interpretive accuracy. Non-expert denotes the crowdsourced majority vote. Gemini-Flash denotes Gemini-3.1-Flash-Lite. ($\dagger$): re-implemented with the Qwen-9B backbone for fair comparison.}

    \begin{tabular}{lllcccc}
        \toprule
        \multirow{2}{*}{\textbf{Group}} & \multirow{2}{*}{\textbf{Method}} & \multirow{2}{*}{\textbf{Backbone}} & \multicolumn{3}{c}{\textbf{Correlation}} & \multicolumn{1}{c}{\textbf{Interpretation}} \\
        \cmidrule(lr){4-6} \cmidrule(lr){7-7}
         &  &  & \textbf{KRCC}~$\uparrow$ & \textbf{SRCC}~$\uparrow$ & \textbf{CCC}~$\uparrow$ & \textbf{Acc (\%)}~$\uparrow$  \\ 
        \midrule
        \multicolumn{7}{l}{\textbf{Conventional Scorers}} \\
 -- & Random Guess & -- & -0.023 & -0.077 & 0.007 & 25.2 \\
Alignment Scoring & CLIPScore~\cite{radford2021learning} & CLIP & 0.041 & 0.059 & 0.106 & -- \\
Quality Scoring & CLIP-IQA~\cite{wang2023exploring} & CLIP & 0.080 & 0.250 & 0.088 & -- \\
Quality Scoring & DeQA-Score~\cite{you2025teaching} & Tuned MLLM & 0.023 & -0.112 & -0.118 & -- \\
 Preference Scoring & Aesthetic Predictor~\cite{schuhmann2022laion} & CLIP & 0.030 & 0.117 & 0.084 & -- \\
 Preference Scoring & PickScore~\cite{kirstain2023pick} & CLIP & 0.202 & 0.605 & 0.310 & -- \\
 Preference Scoring & HPSv2~\cite{wu2023human} & CLIP & 0.030 & -0.017 & -0.081 & -- \\
 Preference Scoring & ImageReward~\cite{xu2023imagereward} & BLIP & -0.002 & 0.232 & 0.128 & -- \\
        \midrule
        \multicolumn{7}{l}{\textbf{Structured-Rationale Evaluators}} \\
Rationale\&Scoring & LMM4LMM~\cite{wang2025lmm4lmm} & Tuned MLLM & 0.274 & 0.651 & 0.547 & 44.0 \\
 Structured Rationale & VIEScore~\cite{ku2024viescore} ($\dagger$) & Zero-shot MLLM & 0.241 & 0.641 & 0.321 & -- \\
 Structured Rationale & DSG~\cite{cho2024davidsonian} ($\dagger$) & Zero-shot MLLM & 0.153 & 0.678 & 0.230 & -- \\
        \midrule
        \multicolumn{7}{l}{\textbf{Art Interpretation / Aesthetic Models}} \\
 Formal Analysis & ArtCoT~\cite{jiang2025multimodal} ($\dagger$) & Zero-shot MLLM & 0.294 & 0.604 & 0.609 & 80.4 \\
 Art Interpretation & ArtiMuse~\cite{cao2025artimuse} & Tuned MLLM & 0.075 & 0.088 & 0.156 & 67.7 \\
 Art Interpretation & GalleryGPT~\cite{bin2024gallerygpt} & Tuned MLLM & -0.034 & -0.201 & -0.087 & 26.1 \\
        \midrule
        \multicolumn{7}{l}{\textbf{Human and SemJudge}} \\
 Human Reference & Human (Non-expert) & -- & 0.790 & 0.924 & 0.946 & 81.5 \\
 Human Reference & Human (Expert) & -- & -- & -- & -- & 93.2 \\
 Semiosis Quality & SemJudge (Qwen-9B) & Zero-shot MLLM & 0.533 & 0.856 & 0.808 & 86.1 \\
 Semiosis Quality & SemJudge (Qwen-35B-A3B) & Zero-shot MLLM & 0.674 & 0.880 & 0.878 & 91.0 \\
        \textcolor{black!60}{Semiosis Quality} & \textcolor{black!60}{SemJudge (Gemini-Flash)} & \textcolor{black!60}{Zero-shot MLLM} & \textcolor{black!60}{0.746} & \textcolor{black!60}{0.964} & \textcolor{black!60}{0.968} & \textcolor{black!60}{92.4} \\
        \bottomrule
    \end{tabular}

    \label{tab:main}
\end{table*}

\emergencystretch=1em

\textbf{Implementation.} We utilize Qwen-3.5-9B as the backbone MLLM unless otherwise specified. This includes all zero-shot MLLM-based baselines for fairness. The MLLM predicts both the HSG schema and the bounding box coordinates in zero-shot (i.e., no finetuning). Each MLLM-based method is evaluated over three runs. For the judgment task, models receive both the user prompt and artifact; for the interpretation task, they receive only the artifact. Additional implementation details, including prompt templates, can be found in Appendix~\ref{appendix:system-prompt}.

\textbf{Compared Methods:} To the best of our knowledge, SemJudge is the first interpretation-centric evaluator for meaning conveyance in HGI, and there are no directly comparable baselines. We therefore compare against three groups of related methods:
\begin{enumerate}
    \item \textbf{Scoring models}, including CLIP-IQA~\cite{wang2023exploring}, DeQA-Score~\cite{you2025teaching}, CLIPScore~\cite{radford2021learning}, PickScore~\cite{kirstain2023pick}, HPSv2~\cite{wu2023human}, ImageReward~\cite{xu2023imagereward}, and LAION Aesthetic Predictor~\cite{schuhmann2022laion}. 
    \item \textbf{Evaluators with structured rationales}: VIEScore~\cite{ku2024viescore},  Davidsonian Scene Graph (DSG)~\cite{cho2024davidsonian}, ArtCoT~\cite{jiang2025multimodal}, and LMM4LMM~\cite{wang2025lmm4lmm}.
        \item \textbf{Art interpretation / aesthetic models}: GalleryGPT~\cite{bin2024gallerygpt} and ArtiMuse~\cite{cao2025artimuse}. %
\end{enumerate} 

\subsection{Correlation with Human Judgments}

\begin{figure*}[!ht]
    \centering
    \includegraphics[width=\linewidth]{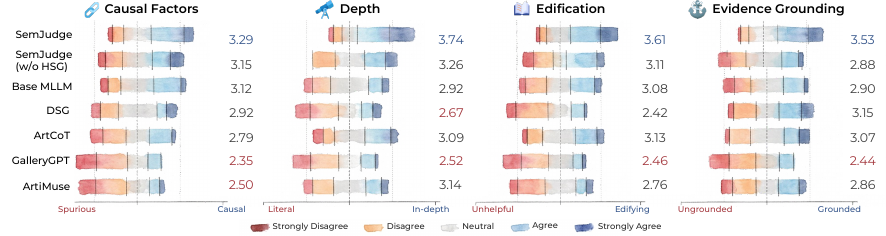}
    \caption{\textbf{Subjective Interpretation Quality Experiment on Four Dimensions.} We show the user ($m=70$) feedback distribution on a 5-point Likert scale, with the mean score shown for each bar. SemJudge (w/o HSG): SemJudge with only root artifact semiosis. Base MLLM: Prompting MLLM to generate art interpretations.}
    \label{fig:fig_user}
\end{figure*}

\textbf{Correlation Metrics.} For statistical alignment analysis, we adopt three complementary metrics capturing different levels of alignment. \textbf{(1) Instance Concordance:} We compute Kendall's tau-b (\textbf{KRCC}) $\tau$ on pairwise 2AFC judgments within each prompt and average over all prompts, measuring concordance with human pairwise preferences at the instance level. \textbf{(2) Discrete Rank Correlation:} Following~\cite{jiang2025multimodal}, we derive Elo scores for the 16 GenArt models from all valid pairwise comparisons and compute Spearman's Rank Correlation Coefficient (\textbf{SRCC}) $\rho$ between the human-derived and metric-derived model rankings. \textbf{(3) Continuous Elo Correlation:} Because SRCC captures only rank order, we additionally report Lin's Concordance Correlation Coefficient (\textbf{CCC}) $\rho_c$ to measure agreement in both ordering and Elo-score magnitude after placing the human- and evaluator-derived Elo scores on a common scale~\cite{lawrence1989concordance}. All metrics lie in $[-1, 1]$, where higher values indicate stronger positive alignment with human judgment.

\textbf{VQA Metric.} For the VQA-based art interpretation task, we report multiple-choice question answering accuracy (\textbf{Acc}), computed as the proportion of correctly answered questions among all questions in the benchmark.

\textbf{Correlation Results.}{\emergencystretch=1em Table~\ref{tab:main} reports agreement with expert judgments from three complementary statistics. Three patterns emerge. \textbf{(1) Conventional scorers show limited human alignment.} Image-quality, prompt-alignment, and preference-based scoring methods exhibit near-zero or weak correlation with expert judgments, suggesting that appearance-level quality and generic preference signals are fundamentally insufficient for evaluating symbolic and indexical meaning conveyance. \textbf{(2) Structured MLLM-based evaluators improve agreement but remain below SemJudge on this benchmark.} Under a shared Qwen-9B backbone, SemJudge outperforms the re-implemented structured baselines, suggesting that the difference cannot be attributed solely to backbone capacity and may also reflect how the evaluation task is formulated for evaluating symbolic and indexical meaning conveyance. }
\textbf{(3) SemJudge achieves the strongest overall human alignment.} With explicit modeling of HGI semiosis through HSGs, SemJudge attains the best performance across all three correlation metrics. This advantage is consistent across different MLLM backbones. These results demonstrate the advantage of semiotics-grounded structured interpretation for human-aligned GenArt evaluation.

\textbf{VQA-based Art Interpretation Results.} Table~\ref{tab:main} also reports the VQA accuracy of the compared methods on fine-grained art interpretation. SemJudge achieves the best overall performance, showing that its semiotics-grounded structure improves not only pairwise judgment alignment but also explicit interpretive understanding. Notably, SemJudge with a lightweight \verb|Gemini-3.1-Flash-lite| achieves a promising 92.4\% accuracy, approaching the expert human performance of 93.2\%.

\subsection{Human Evaluation of Interpretation Quality}\label{sec:interp_qual}

\textbf{Interpretation Quality Dimensions.}
We ask expert and non-expert users to rate generated interpretations on four dimensions: \textbf{Causal Agreement} (experts only), assessing whether the cited decisive factors match the reasons for a 2AFC choice; \textbf{Depth}, assessing meaning beyond literal description; \textbf{Edification for Artwork Comprehension}, assessing whether an interpretation aids understanding of the artwork's intended expression; and \textbf{Evidence Grounding}, assessing support from the prompt and visible image evidence. Each uses a 5-point Likert scale; $4{,}943$ responses were collected. The full questionnaire is provided in Appendix~\ref{app:userstudy}.

\textbf{Discussion.}
Figure~\ref{fig:fig_user} indicates that \textbf{SemJudge} is significantly ($p<0.05$) preferred across all dimensions of subjective interpretation quality. Expert users assign SemJudge the strongest \textbf{Causal Agreement}, suggesting that its decisive factors are better aligned with human reasoning about semiosis quality. SemJudge also receives the highest \textbf{Depth}, consistent with our design goal of interpreting the deep symbolic meaning in HGI. By contrast, the compared methods primarily focus on the appearance of the artifact, such as object presence (DSG) or art style (ArtCoT, ArtiMuse, GalleryGPT). This appearance-focused emphasis is less aligned with object-space meaning conveyance. Users also rate SemJudge highest on \textbf{Edification for Artwork Comprehension}, supporting our motivation that structured semiotic rationales can serve as an accessible bridge between the creator's intent and visual realization and inspire deeper engagement. On \textbf{Evidence Grounding}, both expert and non-expert raters more often judge SemJudge's claims as supported by the prompt and visible evidence, which we attribute to evidence-linked HSG nodes (text spans and bounding boxes) and schemas (Peircean semiosis triad) that help ground interpretation. Appendix~\ref{appendix:exp} provides additional visualizations and qualitative comparisons with the other methods.

\subsection{Empirical Analysis: Iconicity Bias of Conventional Metrics}
\label{sec:icon_bias}

We test whether conventional GenArt evaluators agree with humans primarily on \emph{iconic} prompt--artifact relations.
For each 2-AFC instance $(s^{(1)}, s^{(2)}_a, s^{(2)}_b)$, six human experts rate iconicity, indexicality, and symbolism on 7-point Likert scales.
We combine these into an instance-level \textbf{net iconicity score} $\widetilde{NI}_k$, which is positive when iconic resemblance dominates and negative when symbolic/indexical cues dominate.

\paragraph{Test for iconicity bias.}
For each evaluator and instance $k$, let $\Lambda_k\!=\!1$ if the evaluator's winner matches the human winner (otherwise $\Lambda_k\!=\!0$).
We define:
\[
    \Delta = \mathbb{E}[\widetilde{NI}_k \mid \Lambda_k = 1] - \mathbb{E}[\widetilde{NI}_k \mid \Lambda_k = 0].
\]
A positive $\Delta$ indicates the evaluator aligns with humans mainly on more iconic instances --- an \emph{iconicity bias}.
We assess $H_1\colon \Delta > 0$ via a one-sided permutation test. Full statistical details are in Appendix~\ref{appendix:exp}.

\paragraph{Findings.}
Table~\ref{tab:iconicity-bias} shows that conventional GenArt evaluators exhibit a consistent iconicity bias (significantly $\Delta>0$), which suggests they track human preferences better when artifacts visually resemble their referents.
In contrast, SemJudge’s estimated $\Delta$ is near zero and not statistically significant, suggesting that its agreement with human judgments is not concentrated on the highly iconic subset but can also extend to indexical and symbolic artworks.

\begin{table}[t]
\centering
\small
\setlength{\tabcolsep}{6pt}
\caption{\textbf{Iconicity-bias hypothesis test across evaluators.} We report $\Delta$, bootstrap 95\% confidence intervals, and Cohen's $d$. Sig.\ indicates one-sided permutation-test significance: $^{*}p{<}0.05$, $^{**}p{<}0.01$. Conventional evaluators are biased towards iconic signs, while SemJudge remains robust for symbolic / indexical signs.}
\begin{tabular}{lcccc}
\toprule
Evaluator & $\Delta$ & 95\% CI & Cohen's $d$ & Sig. \\
\midrule
ImageReward   & 0.086 & $[\,0.039,\  \infty)$ & 0.306 & ** \\
PickScore   & 0.126 & $[\,0.047,\  \infty)$ & 0.595 & ** \\
DSG         & 0.087 & $[\,0.006,\  \infty)$ & 0.402 & * \\
ArtCoT      & 0.182 & $[\,0.090,\ \infty)$ & 0.848 & ** \\
\midrule
SemJudge  & -0.010 & $[\,-0.157,\ \infty)$ & -0.047 & \\
\bottomrule
\end{tabular}
\label{tab:iconicity-bias}
\end{table}

\subsection{Ablation Study across MLLMs}
To separate the effect of HSG construction from the underlying MLLM’s raw capability, we organize the ablation around three controlled questions. Table~\ref{tab:additive_ablation} examines: \textbf{(A)} whether adding HSG-based structure improves performance under a fixed judge, \textbf{(B)} whether a high-quality HSG can substantially improve an otherwise lightweight judge, and \textbf{(C)} how much additional benefit is obtained by scaling the final judge once a strong HSG is already available. This design lets us test whether SemJudge's gains arise from structured semiosis reconstruction rather than from backbone scaling alone.

Three findings stand out. \textbf{First}, with the judge fixed, introducing HSG structure improves performance over direct judgment, but weak MLLMs may struggle to faithfully generate highly complex HSGs, which does not always yield further gains. \textbf{Second}, HSGs constructed by a stronger model substantially elevate weak judges, showing that the main bottleneck often lies in HSG construction rather than in the final judge alone. \textbf{Third}, these gains are especially pronounced for VQA, where a strong HSG greatly improves explicit art interpretation. This finding is consistent with the human-based evaluation in Figure~\ref{fig:fig_user}, clearly demonstrating the effectiveness of HSGs for art interpretation.

\begin{table}[t]
\centering
\small
\caption{\textbf{Controlled ablation of SemJudge.} We isolate three questions: (A) whether introducing a standard or more complex HSG structure helps under a fixed judge, (B) whether a strong HSG can lift a weak judge, and (C) how much judge scaling still matters once a strong HSG is available. }
\setlength{\tabcolsep}{5pt}
\resizebox{\linewidth}{!}{
\begin{tabular}{l l l c c}
\toprule
\textbf{HSG Setting} & \textbf{HSG Builder} & \textbf{Judge (2AFC)} & \textbf{KRCC $\uparrow$} & \textbf{VQA Acc $\uparrow$} \\
\midrule
\multicolumn{5}{l}{\textbf{(A) Same judge, vary HSG complexity}}\\
No HSG     & --              & Qwen-9B          & 0.48 & 82.0 \\
Standard HSG & Qwen-9B         & Qwen-9B          & 0.55 \UpUpG & 86.1 \UpG \\
Complex HSG   & Qwen-9B         & Qwen-9B          & 0.51 \UpG & 84.3 \UpG \\
\midrule
\multicolumn{5}{l}{\textbf{(B) Strong HSGs can lift weak judges}}\\
No HSG     & --              & Qwen-2B          & -0.04 & 24.1 \\
No HSG     & --              & Qwen-4B          & 0.28 & 56.8 \\
Complex HSG   & Gemini-Flash    & Qwen-2B          & 0.27 \UpUpG & 42.2 \UpUpG \\
Complex HSG   & Gemini-Flash    & Qwen-4B          & 0.52 \UpUpG & 86.8 \UpUpG \\
\midrule
\multicolumn{5}{l}{\textbf{(C) Residual effect of judge scaling with the same strong HSG}}\\
Complex HSG   & Gemini-Flash    & Qwen-9B          & 0.57 \UpUpG & 91.6 \UpUpG \\
Complex HSG   & Gemini-Flash    & Gemini-Flash     & 0.73 \UpUpG & 92.4 \UpUpG \\
\bottomrule
\end{tabular}}
\label{tab:additive_ablation}
\end{table}

\subsection{Limitations}
SemiosisArt leverages Christian, East Asian, Hindu, and Islamic traditions and modern artistic motifs as anchors for art interpretation. While this provides a more culturally grounded and inter-subjective approach to reliable benchmarking, we acknowledge that it may not fully represent the diversity of artistic expression or the associated interpretive challenges in generative art. Minority cultural traditions and contemporary conceptual art are two major categories that may not be well represented. This is mainly because stable shared human judgments for these relatively niche domains are difficult to obtain, both in principle~\cite{eco1989open} and in our study. We extend this discussion in Appendix~\ref{app:discussion} and hope that future work can further address these limitations.

\section{Conclusion}

Our study highlights an important aspect that current GenArt evaluators do not adequately capture: the symbolic and indexical dimensions of visual art. As artistic practice can extend beyond perceptual resemblance toward conceptual meaning, GenArt evaluation may also benefit from attending to the ideas and intentions conveyed through generated works. By integrating semiotic theory, we broaden the evaluative focus from surface-level appearance to processes of meaning-making. Our findings suggest that existing evaluators tend to align more strongly with iconic resemblance, whereas SemJudge more explicitly models the interpretive process through which meaning is conveyed from creators' intentions to generated artifacts. This leads to improved alignment with human judgments and subjective interpretation quality. We hope this work encourages further research on the interpretive dimensions of GenArt and their role in artistic evaluation.

\bibliographystyle{ACM-Reference-Format}
\bibliography{main}

\clearpage
\appendix

\section{Details of SemiosisArt}
\label{appendix:dataset}
Constructing a meaning-oriented GenArt dataset is critical for semiosis quality evaluation. This section provides details on how SemiosisArt is constructed to focus on meaning and interpretation, instead of appearance, as in existing datasets.

\paragraph{Dataset Construction.} We construct the dataset iteratively with expert feedback and crowdsourced quality control. In brief, expert users propose prompts whose intended meanings rely substantially on symbolic or indexical interpretation while remaining sufficiently canonically grounded to support inter-subjective judgment from non-expert users. In practice, this means anchoring prompts to theological stories, literature, painting traditions, cultural motifs, and other established interpretive sources, rather than relying on unconstrained free-form readings.

\begin{figure}[!ht]
    \centering
    \resizebox{\linewidth}{!}{%
        \includegraphics{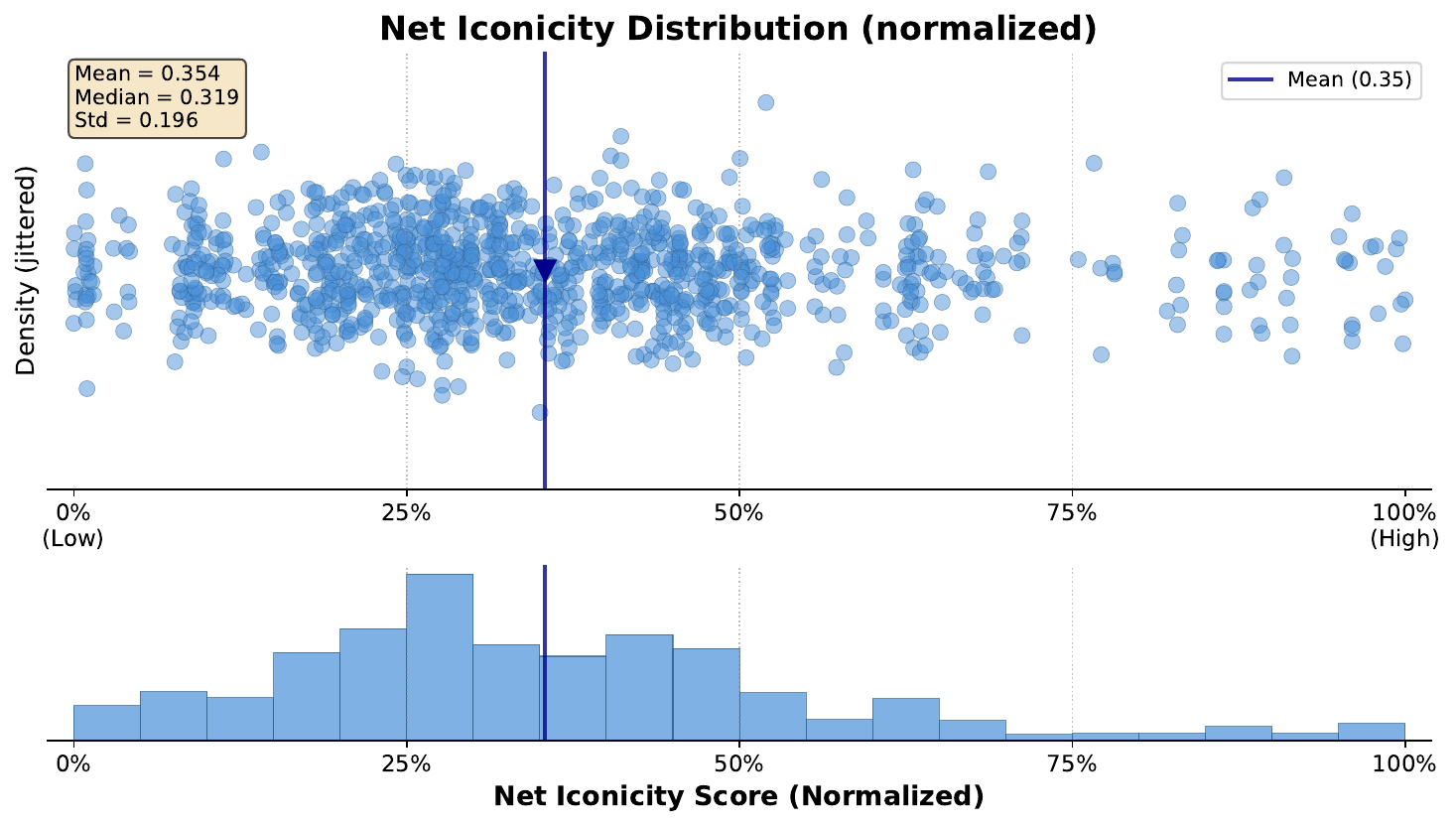}
 }
    \caption{Net iconicity distribution of SemiosisArt, jittered and normalized, with outliers omitted for plotting.}
    \label{fig:icon_dist}
\end{figure}

\paragraph{Scale.} SemiosisArt contains $187$ HSG instances and outputs from a pool of $16$ generative models. For each instance (i.e., input prompt), we sample $5$ models to generate $5$ images, yielding $187\times 5 = 935$ images in total. These $5$ images induce $\binom{5}{2}=10$ pairwise comparisons per instance, so the judgment task contains $187\times \binom{5}{2}=1{,}870$ 2AFC comparative judgment instances. In addition, the benchmark includes $600$ VQA questions for fine-grained interpretation. We construct the benchmark with $m_1=12$ experts and use $38{,}155$ non-expert judgments for quality control, retaining tasks with sufficient inter-subjective agreement for evaluation. We chose this scale to balance (i) coverage across diverse motifs and traditions, (ii) statistical power for correlation analyses, and (iii) the practical cost of repeatedly querying API-based MLLMs across all benchmark instances. In particular, scaling this expert-annotated, art-centric dataset is especially costly because it requires sustained interaction with experts from different cultural backgrounds, each contributing culturally grounded judgments and fine-grained annotations. These annotations go beyond simple 2AFC labeling and require an iterative quality-control process, detailed below, to ensure that symbolic, stylistic, and contextual interpretations are accurately captured. This annotation burden partly explains the relatively modest size of the dataset.

\paragraph{Iterative Process.} The dataset construction process is iterative, with multiple rounds of expert feedback and crowdsourced quality control. In each round, experts propose challenging (i.e., low-iconicity) prompts, which are then used to generate images from the pool of models. These images are subjected to crowdsourced 2AFC judgments, and we retain only prompts that yield sufficient inter-subjective agreement. In practice, we first filter out prompts with less than 60\% effective agreement at the instance level and label the corresponding 2AFC tasks as unreliable. We then filter out instances with fewer than four reliable 2AFC tasks to avoid a highly sparse tournament graph, although this is not a sufficient condition. At the end of each round, experts revise prompts associated with unreliable 2AFC tasks, and we regenerate the images. We repeat this process for three rounds. Each 2AFC task is judged by at least 13 non-expert users.

\paragraph{VQA Generation Process.} We generate VQA questions through a semi-automatic process. For each instance, we select the top-ranked image among the five generated images according to Elo ratings from human judgments. We then ask experts to annotate regions of interest in the image. These annotations correspond to symbolic or indexical objects in the image, akin to sub-semioses in HSGs. We then use MLLMs for question generation. Specifically, we use GPT-5.4 to propose questions across 10 question types, followed by quality control with Gemini-3-Pro. This process filters out questions that are either too easy (e.g., answerable from surface-level cues or language alone) or too difficult (e.g., requiring esoteric knowledge or highly subjective interpretation). We also use a held-out set of two experts ($m_3=2$) for quality control under the same standard as the automatic filtering. If a question fails the quality check, we regenerate it. The instruction details are summarized in Table~\ref{tab:qa_inst}. The final VQA questions pass both automatic filtering and expert review, ensuring that they are appropriately challenging and relevant for evaluating semiosis quality. Expert VQA labels are determined by the majority vote of the remaining $12-2=10$ experts.

\paragraph{Generation models.}
The benchmark is constructed from a pool of $16$ contemporary GenArt systems spanning commercial and open-generation/editing models. These include GPT-Image-1.5~\cite{openai_gptimage15_2026}, GPT-Image-1-Mini~\cite{openai_gptimage1_2025}, Nano-Banana-Pro~\cite{nanobanana_2026}, Nano-Banana~\cite{nanobanana_2026}, 
Nano-Banana-2~\cite{nanobanana_2026}, 
Kling-Image-O1~\cite{klingteam2025klingomni},
Grok-Imagine-Image~\cite{grok_imagine},
Qwen-Image-2.0~\cite{qwen_image2},
SeedDream-4.0~\cite{bytedance_seedream4_2025}, Dreamina-3.1~\cite{dreamina2024}, Z-image~\cite{cai2025z}, Qwen-Image-20B~\cite{wu2025qwen}, Qwen-Edit~\cite{wu2025qwen}, Imagen-3-Fast~\cite{baldridge2024imagen}, Ideogram-v2a-Turbo~\cite{ideogram2024}, and Flux-2-Dev~\cite{flux2024}.

\paragraph{Raw Elo Score (Human Labeling)} 

Below, we report an illustrative Elo ranking from our human study for a subset of benchmarked generation models. We set the initial rating as $1{,}500$ with $k=32$:

\begin{verbatim}
 -------------
 1. nano-banana-2                      : 1926.32
 2. Nano-Banana-Pro                    : 1859.83
 3. GPT-Image-1.5                      : 1827.35
 4. Grok-Imagine-Image                 : 1677.60
 5. Seedream-4.0                       : 1637.95
 6. Nano-Banana                        : 1611.65
 7. Qwen-Image-2.0                     : 1589.55
 8. Kling-Image-O1                     : 1573.33
 9. GPT-Image-1-Mini                   : 1556.89
 10. FLUX-2-Dev                         : 1433.02
 11. Dreamina-3.1                       : 1369.26
 12. Qwen-Image-20B                     : 1323.84
 13. Qwen-Edit                          : 1285.95
 14. Z-Image-Lightning                  : 1237.26
 15. Imagen-3-Fast                      : 1076.80
 16. Ideogram-v2a-Turbo                 : 1013.40
 -------------
\end{verbatim}

\begin{figure*}[!ht]
    \centering
    \resizebox{\linewidth}{!}{%
        \includegraphics{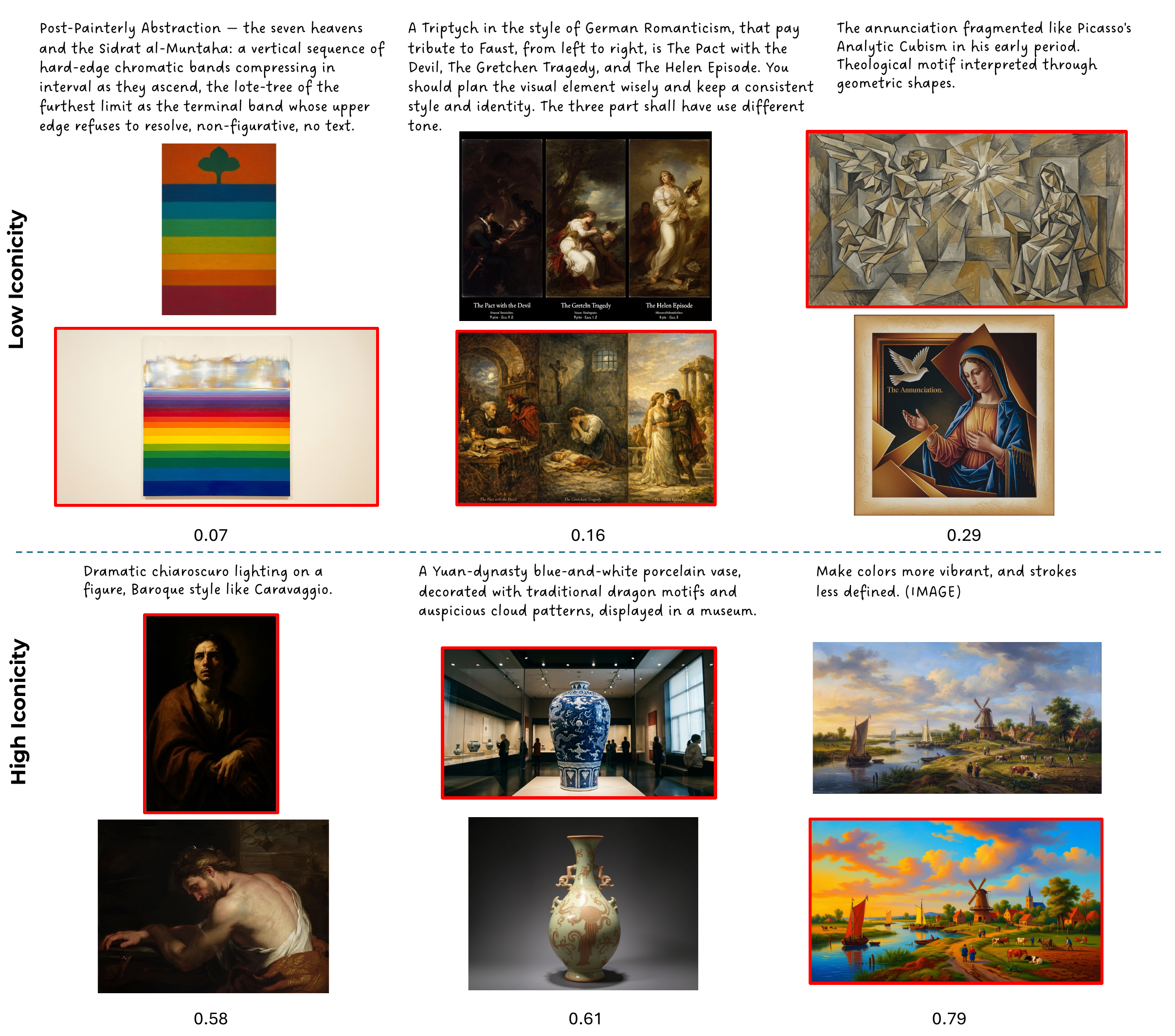}
 }
        \caption{\textbf{2AFC tasks (prompt, image pair) with net-iconicity annotations.} The image with a red border indicates the winner in human annotation. This designation does not represent the iconicity of the image itself.}
    \label{fig:dataset_sample}
\end{figure*}

\paragraph{Visualizations}
In Figure~\ref{fig:dataset_sample}, we visualize images annotated with normalized net-iconicity ($NI$) scores. Figure~\ref{fig:icon_dist} shows the distribution of net-iconicity scores. High-iconicity tasks mostly focus on low-level features, such as tone, or on reference images for identity preservation. Low-iconicity tasks relate to history, convention, stories (e.g., theological narratives), and causality (e.g., mirror reflections). A random sample of 15 prompts is provided in Table~\ref{tab:prompt_sample}.

\begin{table*}[!h]
\caption{Random sample of 15 prompts from the SemiosisArt.}
\centering
\scriptsize
\begin{tabular}{p{1.6\columnwidth} p{0.25\columnwidth}}
\toprule
\textbf{User Prompt (After translation, if necessary)} & \textbf{Iconicity / Symbolism / Indexicality} \\
\midrule
Portrait of Power from Chainsaw Man, wearing a casual oversized sweater, making a peace sign gesture, anime style. &
6.4 / 3.4 / 1.8 \\
\midrule
HD graphite drawing of a robotic hand holding a reflective metal sphere in first person perspective. The ball reflects the android's body, the surrounding warehouse environment. &
4.0 / 3.0 / 5.8 \\
\midrule
The Garden of Earthly Delights by Hieronymus Bosch, but the three panels use different art styles. Left panel use the common style in Renaissance. The center panel use Fauvism by Henry Martisse, and right one use Picasso's Cubism. The story should be still recognizable. Keep the structure of Triptych. &
5.6 / 5.8 / 2.0 \\
\midrule
It turns out that long ago, when Nüwa refined stones to mend the sky, she forged at the Crag of Inaction on Mount Great Desolation a total of 36,501 stones—each twelve zhang high and twenty‑four zhang wide. Of these, the Divine Empress Nüwa used 36,500 stones, leaving one single stone unused, which she discarded beneath Green Ridge Peak of this mountain. Who would have thought that after undergoing refinement, this stone had already gained spiritual awareness? Seeing that all the other stones were chosen to repair the heavens while it alone was deemed unfit and left behind, it fell into deep self‑pity and lamentation, crying out in shame and sorrow day and night.' Based on the above story, create a traditional Chinese lianhuanhua (illustrated serial comic). The narrative should be complete yet concise. &
3.6 / 5.4 / 2.2 \\
\midrule
The Return of the Prodigal Son happened in Minecraft world. The story should be recognizable. &
4.8 / 4.6 / 1.8 \\
\midrule
Create an illustrated step-by-step diagram showing how this sculpture is made in five stages, starting from a solid marble cube and ending with the finished sculpture displayed in a gallery. The diagram should be clear, simple, and easy for a 12-year-old to understand. \verb|[IMAGE]| &
5.8 / 3.0 / 5.2 \\
\midrule
Renaissance art seeks timeless harmony; Baroque art dramatizes the moment in motion. Using the theme of Arrest of Jesus, give a side-by-side comparison to visually demonstrate this difference (generate an image), beyond their appearance-level difference. &
3.2 / 5.8 / 4.0 \\
\midrule
The magnificent scene conveyed in meaning in the Tears in Rain monologue (also known as the C‑Beams Speech), rather than a literal depiction of a man standing in the rain. &
3.6 / 6.0 / 2.2 \\
\midrule
Generate a painting of Marcel Duchamp in the same style as the provided style reference. Do not simply copy the elements from reference but adapt Duchamp's features into that style. \verb|[STYLE_REFERENCE_IMAGE]| &
5.4 / 4.2 / 2.0 \\
\midrule
Interpret the core ideas explored in Zhuangzi’s Qiwulun (On the Equality of Things) using a traditional Chinese ink painting approach, integrating both calligraphy and pictorial elements. Execute the work with a dry‑brush technique on paper. &
3.8 / 6.4 / 3.0 \\
\midrule
Ottoman Iznik ceramic tile art aesthetic for the door-knocking parable in Rumi's Masnavi — the man who knocks on a beloved's door and, asked 'Who is there?', answers 'I', is turned away; returns after years of burning in love, knocks again, is asked 'Who is there?', answers 'You', and is immediately welcomed — characteristic Iznik cobalt, turquoise, and Armenian red on white ground with tulip and saz leaf surrounds framing a two-register narrative of separation and union, no text.&
2.8 / 6.0 / 5.3 \\
\midrule
Storyboard of the Honnō-ji Incident (use 4 panels) rendered in 1980s VHS style&
3.8 / 5.0 / 5.2 \\\midrule
Odysseus and the Sirens as a maritime safety warning sign. The graphic design should follow modern ISO safety signage conventions. &
3.8 / 6.8 / 3.0 \\\midrule
Haru wa akebono. Yauyau shiroku nariyuku yamagiwa, sukoshi akarite, murasaki-dachitaru kumo no hosoku tanabikitru. Natsu wa yoru. Tsuki no koro wa saranari, yami mo naho, hotaru no oku tobichigaitaru. Mata, tada hitotsu futatsu nado, honoka ni uchihikarite yuku mo okashi. Ame nado furu mo okashi. Aki wa yugure. Yuuhi no sashite yama no ha ito chikau naritaru ni, karasu no nedokoro e yuku tote, mitsu yotsu, futatsu mitsu nado tobiiosogu sae aware nari. Maite kari nado no tsuranetaru ga, ito chiisaku miyuru wa, ito okashi. Hi iri hatete, kaze no oto, mushi no ne nado, hata iu beki ni arazu. Fuyu wa tsutomete. Yuki no furitaru wa iu beki ni mo arazu, shimo no ito shiroki mo, mata sarademo ito samuki ni, hi nado isogi okoshite, sumi mote wataru mo, ito tsukizukishi. Hiru ni narite, nuruku yurubimote ikeba, hibachi no hi mo, shiroki hai-gachi ni narite waroshi. Reinterprete these scenes in outsider art (Art Brut) style. Avoid rendering texts. &
4.2 / 5.8 / 3.2 \\
\midrule
modern vector art illustration style for Farid al-Din Attar's Ilahinama (Book of God), respecting the classical symbolism. &
3.2 / 6.2 / 3.2 \\
\bottomrule
\end{tabular}

\label{tab:prompt_sample}
\end{table*}

\section{Additional Experiment Results}

\label{appendix:exp}
\subsection{Bounding Box Grounding Quality}
This section provides additional details about bounding-box prediction. 
\emergencystretch=1em

\paragraph{Motivation for reporting satisfaction rate.} 
Standard detection metrics, such as Mean Intersection over Union (mIoU), rely on fixed ontologies and pre-annotated ground-truth masks (e.g., COCO or LVIS). However, SemJudge operates in an \textbf{open-vocabulary, generative setting}. Since the model dynamically generates the label (the sub-sign description) based on its interpretation of the artwork, no static ground truth exists for these generated concepts. Moreover, the spatial spans of art-related symbols can be vague and open to interpretation, making it difficult to define a single ``correct'' bounding box. Consequently, calculating mIoU against a static baseline is mathematically ill-posed in this context.

\paragraph{Method design.}
Rather than expecting an exact box match, we evaluate whether the predicted box is \textbf{pragmatically useful} for semiotic analysis. Concretely, we report a \textbf{human satisfaction rate} that measures whether the predicted box provides valid visual evidence for the semiotic argument constructed by the model.

\paragraph{User study.}
We measure bounding-box quality through a user study. Specifically, we present the image together with its bounding-box visualization and ask annotators whether the visualization is satisfactory using a binary choice. We collected 450 satisfaction annotations. 

Among all models, Gemini-3.1-Flash-Lite has the highest satisfaction rate (74.7\%), compared with Qwen-3.5-35B-A3B (56.3\%) and Qwen-3.5-9B (52.7\%). This is consistent with their correlation and VQA performance. 

\paragraph{Limitations and Future Work.} MLLMs are known to be limited in directly predicting precise bounding boxes in zero-shot settings~\cite{liu2025can,ma2024groma}. A dedicated grounding module could provide stronger localization capability and potentially higher satisfaction rates. Models such as GroundingDINO~\cite{liu2024grounding}, which can also operate zero-shot, are promising candidates for this purpose. We view such a module as an incremental extension to our theoretical framework and therefore do not include it as part of our contribution.

\subsection{Details of the Iconicity-Bias Analysis}

We test whether conventional GenArt evaluators agree with humans primarily on \emph{iconic} prompt--artifact relations. To do so, we first quantify a \textbf{subjective iconicity score} for each benchmark instance.

\paragraph{Net iconicity score.}
For each 2AFC instance with $(s^{(1)}, s^{(2)}_a, s^{(2)}_b)$, we estimate how much the judgment is driven by iconic resemblance versus indexical or symbolic cues. Because a sign may simultaneously contain all three components, we define the net iconicity score of a sign as
\[
NI(s)= Icn(s) - \tfrac{1}{2}\big(Idx(s)+Sym(s)\big),
\]
where $Icn$, $Idx$, and $Sym$ are 7-point Likert ratings of iconicity, indexicality, and symbolism, respectively, provided by six human experts.

We then aggregate sign-level scores into an instance-level score:
\[
\widetilde{NI}_k
= NI\!\big(s^{(1)}_k\big)
+ \tfrac{1}{2}\Big(NI\!\big(s^{(2)}_{k,a}\big) + NI\!\big(s^{(2)}_{k,b}\big)\Big).
\]
Positive $\widetilde{NI}_k$ indicates that the instance is dominated by iconic resemblance, while negative values indicate a stronger role for symbolic and indexical cues.

\paragraph{Hypothesis test.}
For each evaluator and instance $k$, let $\Lambda_k=1$ if the evaluator's winner matches the human winner, and $\Lambda_k=0$ otherwise.
We compare the average iconicity of the aligned and misaligned subsets:
\[
\Delta=\mathbb{E}[\widetilde{NI}_k\mid \Lambda_k=1]-\mathbb{E}[\widetilde{NI}_k\mid \Lambda_k=0].
\]
A positive $\Delta$ indicates that the evaluator agrees with humans mainly on more iconic instances, which we interpret as an \emph{iconicity bias}.
We test the one-sided hypothesis $H_1\colon \Delta>0$ using a permutation test, and report a one-sided 95\% bootstrap confidence interval together with Cohen's $d$ as the effect size.

\paragraph{Interpretation.}
Under this setup, an evaluator with a strong resemblance bias should align with humans more often on highly iconic cases than on symbolic or indexical cases, producing $\Delta>0$.
By contrast, an evaluator that is robust across different semiotic regimes should not concentrate its agreement on the iconic subset, and therefore should not exhibit a significantly positive $\Delta$.

\subsection{Additional Visualization of HSGs}

\begin{figure*}
    \centering 
    \resizebox{\linewidth}{!}{    \includegraphics{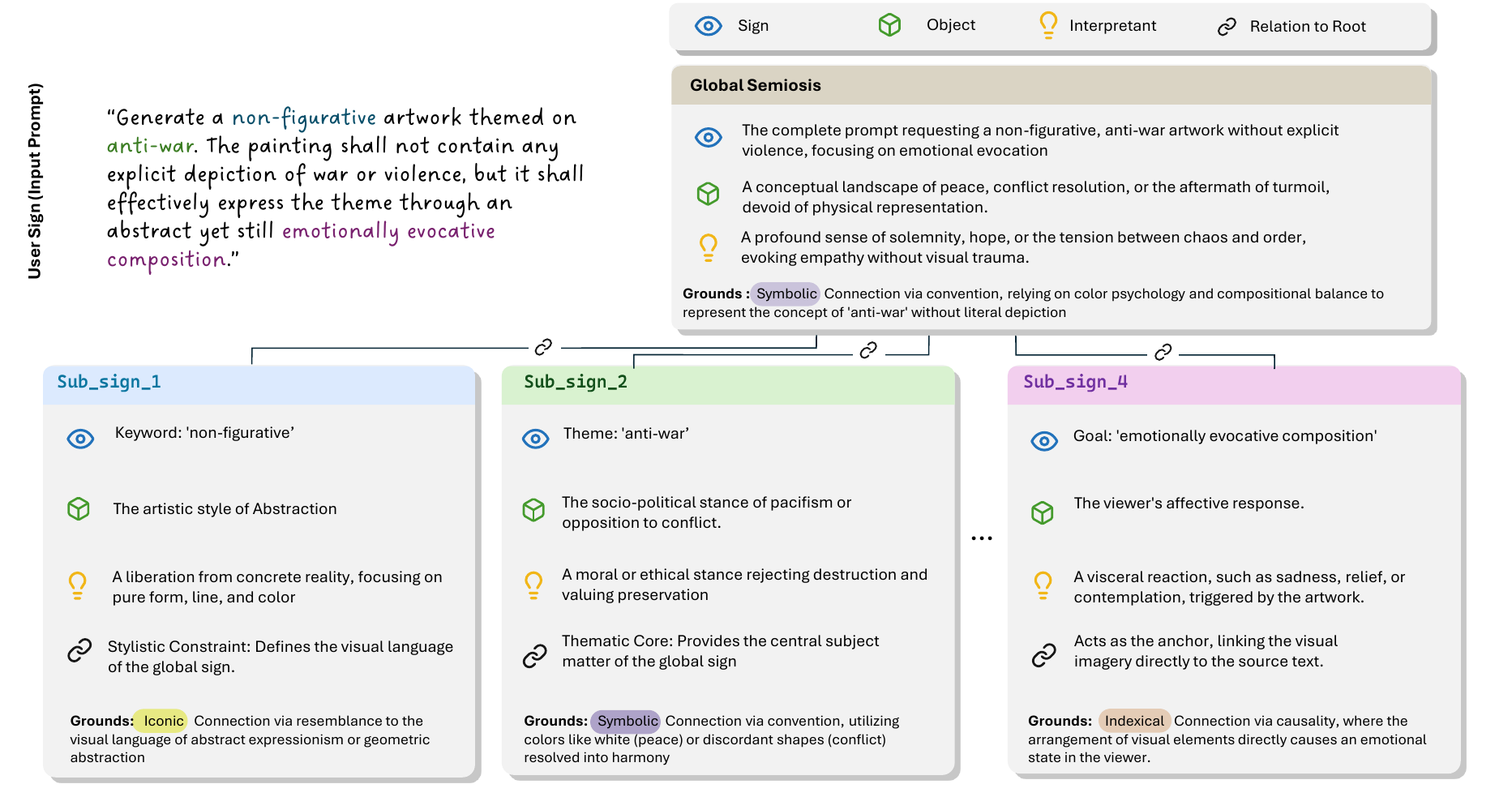}}

    \caption{\textbf{HSG Visualization for User Sign.} Best viewed in color.}
    \label{fig:more_hsg_user}
\end{figure*}

\begin{figure*}
    \centering 
    \resizebox{\linewidth}{!}{    \includegraphics{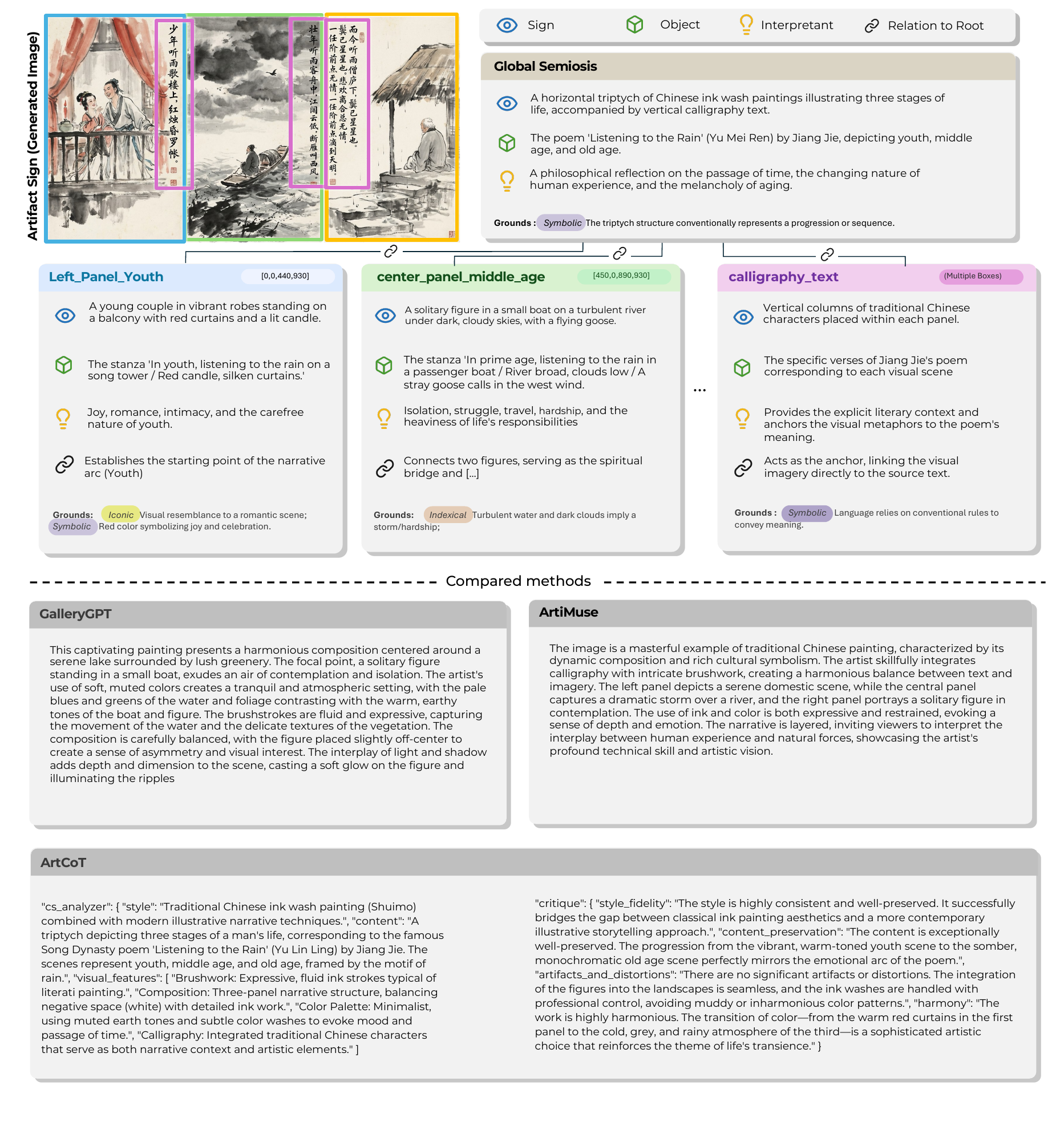}}

    \caption{\textbf{HSG Visualization for Artifact Sign 1.} Best viewed in color. The prompt associated with the image is: Create an artistic-conception illustration inspired by Jiang Jie’s ``Yu Meiren · Listening to the Rain'' in the style of freehand ink-wash painting, using traditional Chinese artistic techniques to highlight the contrasts expressed in the poem. Top: SemJudge output HSG. Bottom: Art analyses from compared models.}
    \label{fig:more_hsg_1}
\end{figure*}

\begin{figure*}
    \centering 
    \resizebox{\linewidth}{!}{    \includegraphics{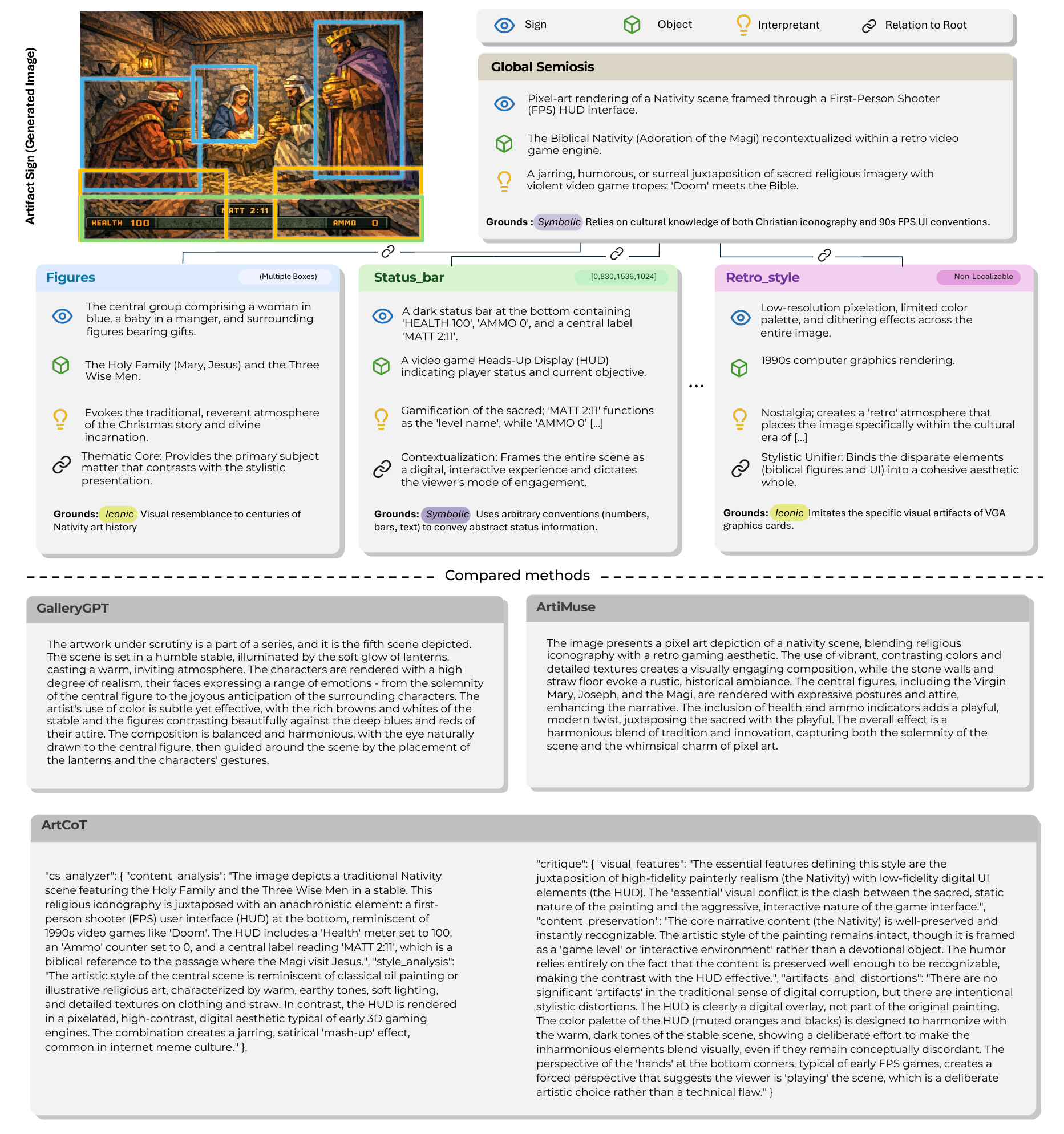}}

    \caption{\textbf{HSG Visualization for Artifact Sign 2.} Best viewed in color. The prompt associated with the image is: Render the story of Matthew 2:11 in a millennial-era video game style, such as Half-Life 1. The clothing and environment should still reflect the historical period. Top: SemJudge output HSG. Bottom: Art analyses from compared models.}
    \label{fig:more_hsg_2}
\end{figure*}

\begin{figure*}
    \centering 
    \resizebox{\linewidth}{!}{    \includegraphics{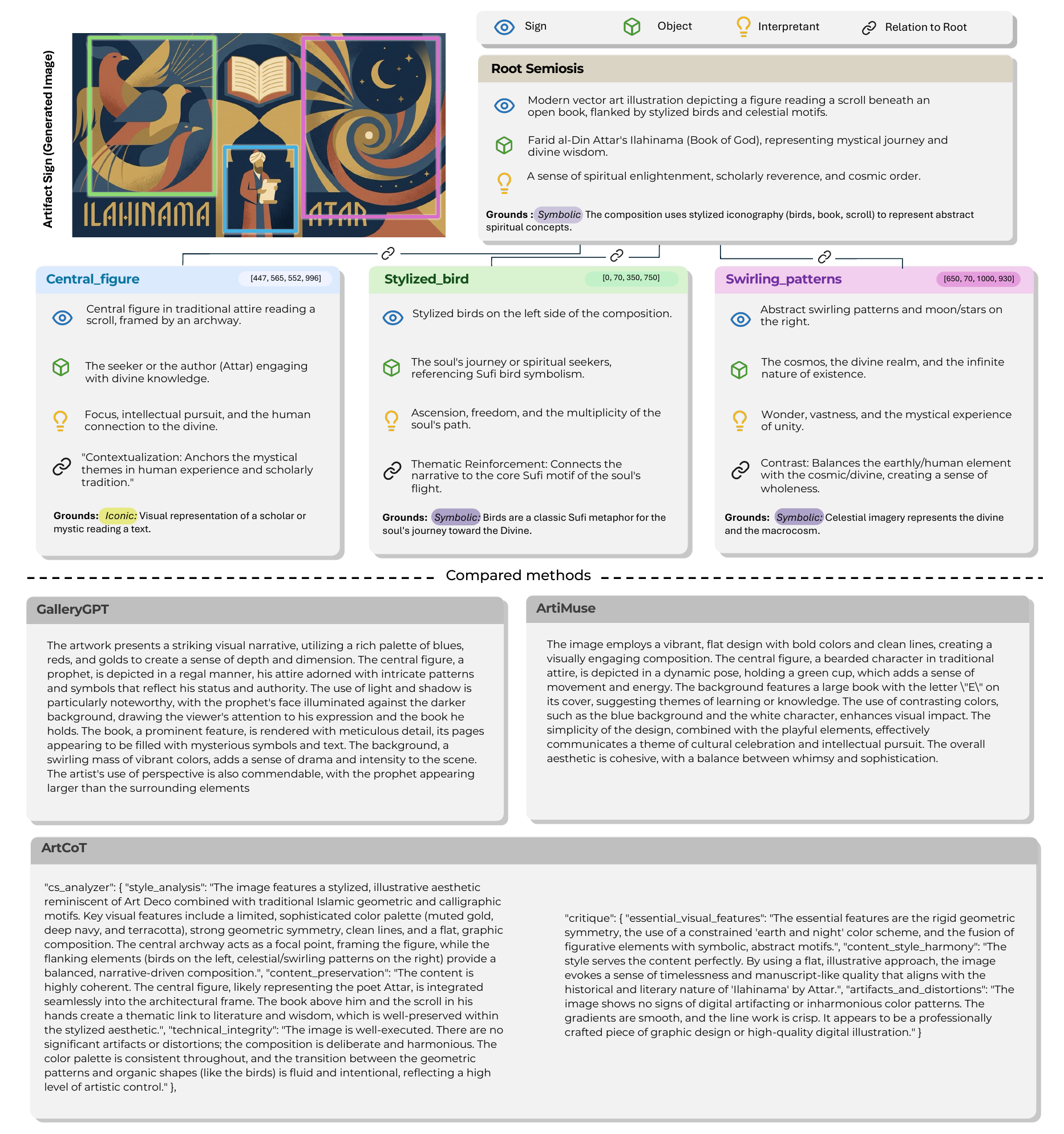}}

    \caption{\textbf{HSG Visualization for Artifact Sign 3.} Best viewed in color. The prompt associated with the image is: Modern vector art illustration style for Farid al-Din Attar's Ilahinama (Book of God), respecting the classical symbolism. Top: SemJudge output HSG. Bottom: Art analyses from compared models.}
    \label{fig:more_hsg_3}
\end{figure*}

Figure~\ref{fig:more_hsg_user} provides a visualization of the user sign (prompt), and Figures~\ref{fig:more_hsg_1}, \ref{fig:more_hsg_2}, and \ref{fig:more_hsg_3} provide three additional HSG visualizations for artifact signs.

\begin{figure*}
    \centering 
    \resizebox{\linewidth}{!}{    \includegraphics{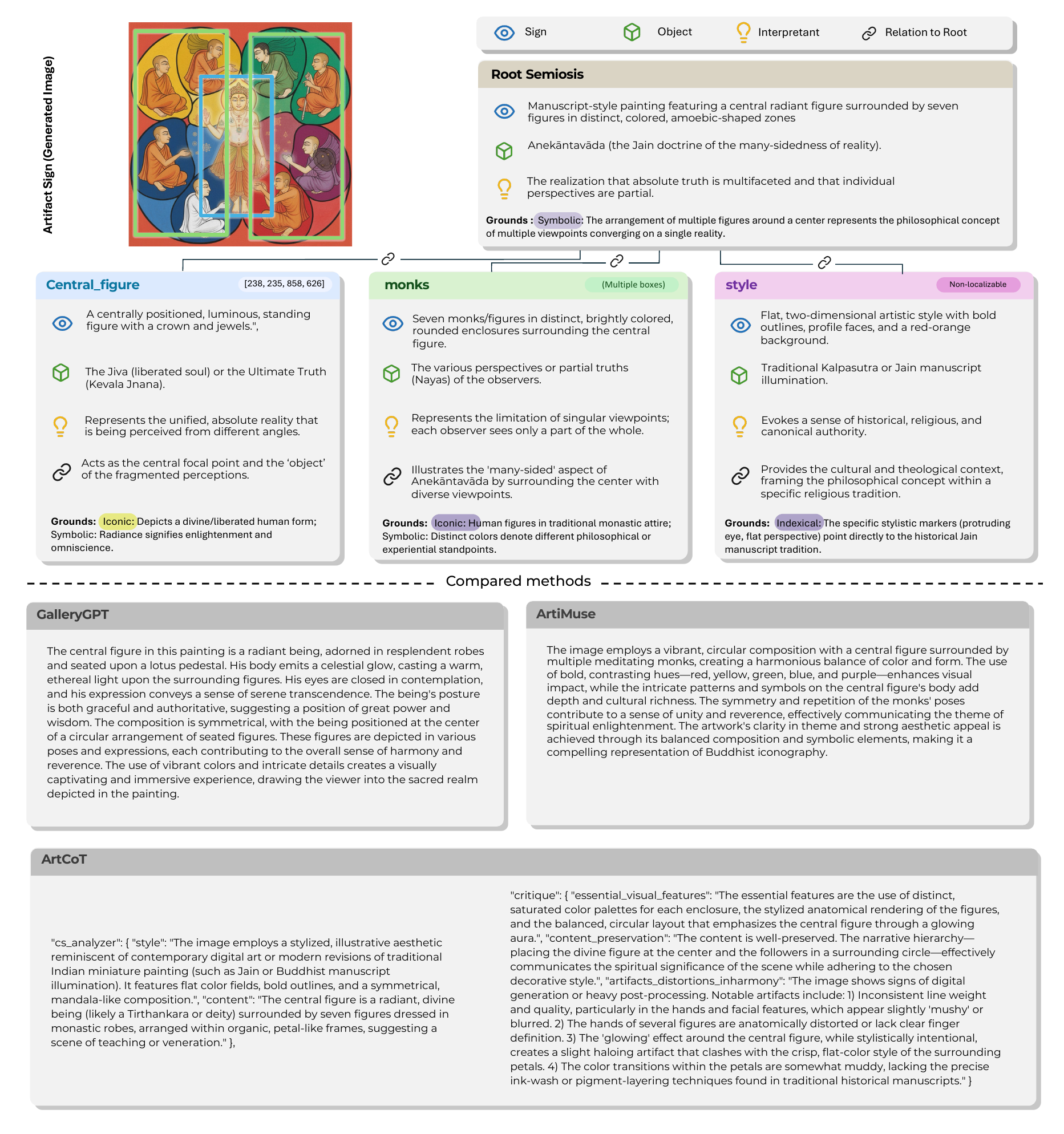}}

    \caption{\textbf{HSG Visualization for Artifact Sign 4.} Best viewed in color. The prompt associated with the image is: Jain manuscript painting style (in the tradition of Kalpasutra illustrations) depicting the philosophical concept of Anekāntavāda (the many-sidedness of truth) — the parable of the blind men and the elephant reimagined with Jain symbolic figures in distinct colored zones, each perceiving a fragment of a radiant liberated Jiva, with characteristic red-orange ground and flat-profile faces and no text. Top: SemJudge output HSG. Bottom: Art analyses from compared models.}
    \label{fig:more_hsg_4}
\end{figure*}

\section{User Study Detail}
\label{app:userstudy}
\subsection{Raw Annotation Acquisition}
\paragraph{Setup.}
We ran a human study to obtain meaning-level judgments for prompted generative art and to validate evaluator alignment with human preferences.
The study includes both \textbf{2AFC} judgments and \textbf{VQA} annotations. In each 2AFC trial, participants view a prompt and two generated images, then select the image that better conveys the prompt's intended meaning, taking into account metaphor, symbolism, convention, and other meaning-level cues rather than only literal object matching. In the VQA setting, participants answer or validate fine-grained interpretive questions about the prompt--image pair, focusing on whether specific symbolic, contextual, or meaning-level relations are correctly recognized.

\paragraph{Participants and task distribution.}
We include both expert and non-expert participants in this study.
\begin{enumerate}
    \item Expert participants ($m_1=12$) have formal training in art-related disciplines, or relevant studio experience. They are responsible for initiating and refining canonically grounded prompts with substantial symbolic or indexical meaning. They also annotate 2AFC tasks, contribute to VQA construction and/or verification, and participate in the fine-grained interpretation-quality study.
    \item Non-expert participants are recruited from crowdsourcing platforms. We apply quality control by monitoring completion behavior and replacing invalid responses when necessary. For the interpretation-quality study reported in the main paper, expert and non-expert participants together contribute a total of $m=70$ raters and $4{,}943$ Likert-scale responses. 
\end{enumerate}
Experts are required to annotate all 2AFC comparison tasks. For non-expert users, we randomly sample a subset of tasks for each participant to control study duration. The VQA set is constructed semi-automatically: experts first subsample SemiosisArt instances according to their domain expertise, then GPT-5.4 proposes canonical questions for these prompt-image pairs, and Gemini-3-Pro subsequently audits or rewrites the drafted questions for clarity and canonical faithfulness. Human experts then review the resulting candidates and reject malformed or weak questions until a final pool of $600$ questions is obtained. The sequence of task presentation and the left/right candidates are randomized in all settings. GenArt model names are hidden from participants during annotation.

\begin{figure*}
    \centering
    \includegraphics[width=\linewidth]{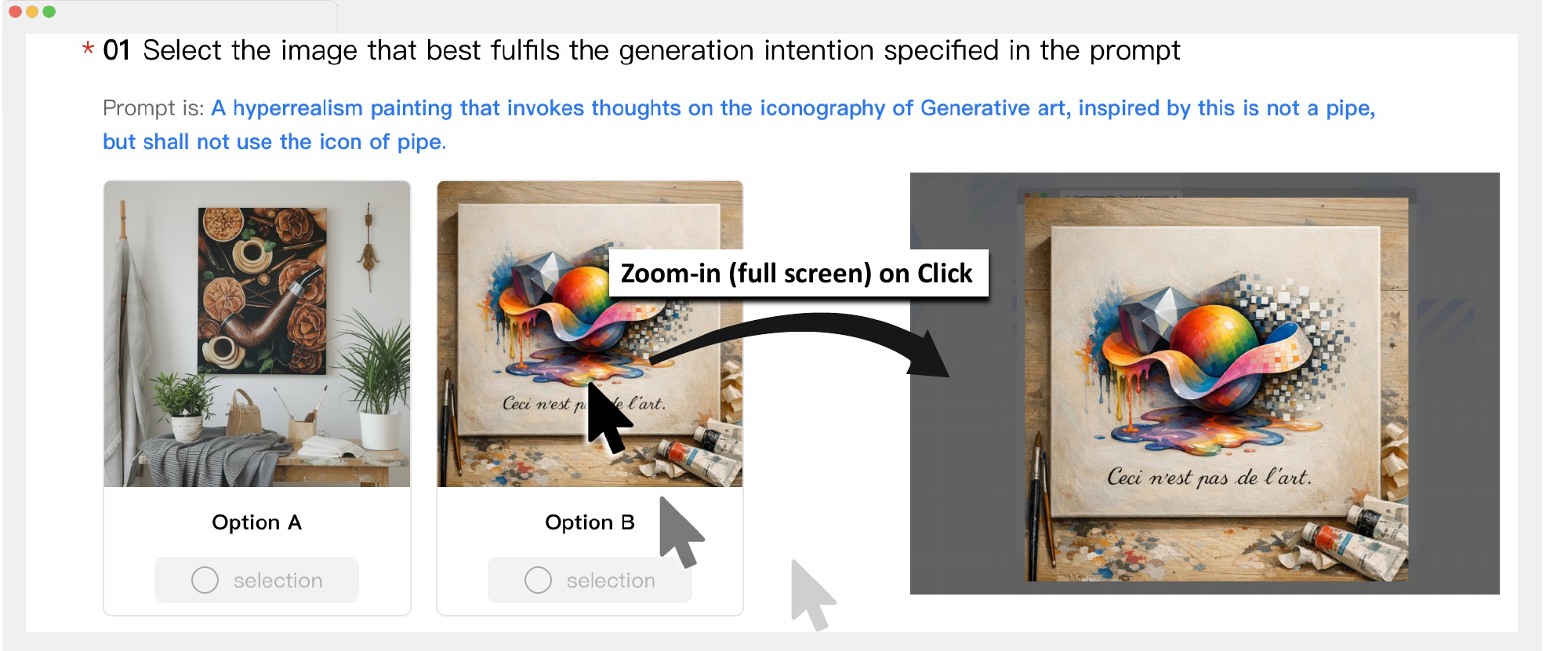}
    \caption{2AFC User Annotation Interface. Participants must choose the better image in a pairwise comparison. The input prompt is provided, and an image can be enlarged by clicking its option card.}
    \label{fig:user_2afc}
\end{figure*}

\begin{figure}
    \centering
    \includegraphics[width=\linewidth]{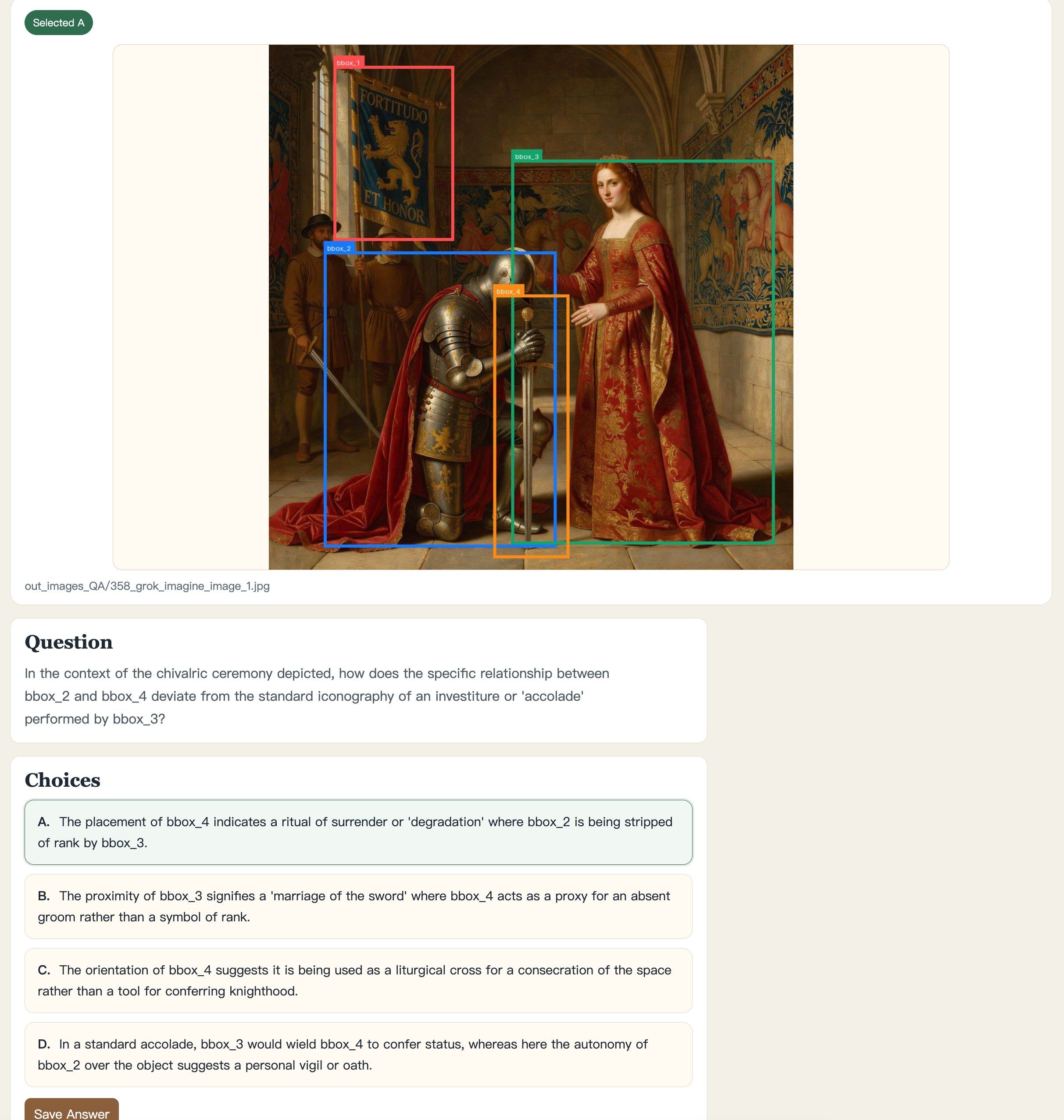}
    \caption{VQA User Annotation Interface.}
    \label{fig:user_vqa}
\end{figure}

\begin{figure*}[!ht]
    \centering
    \includegraphics[width=\linewidth]{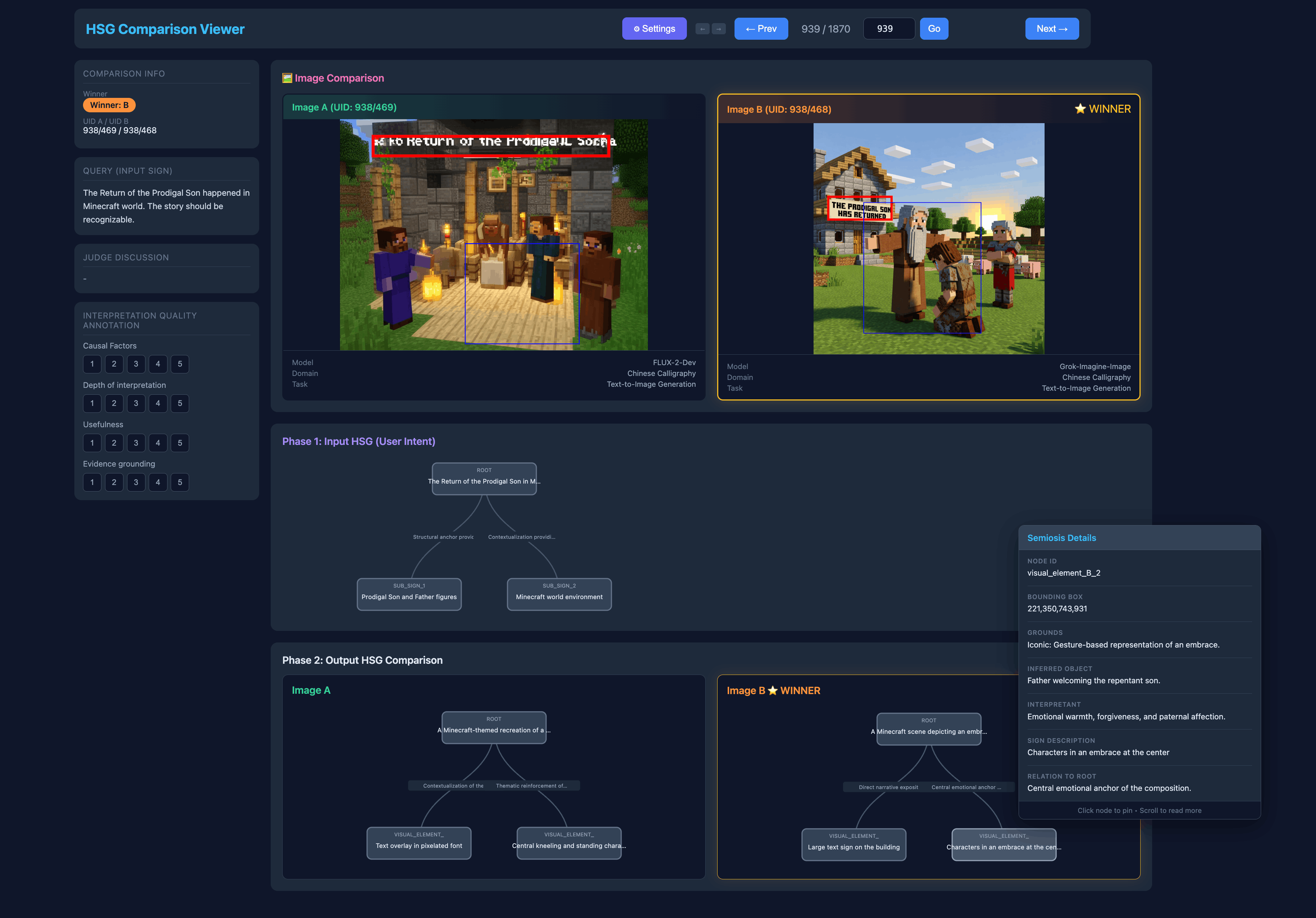}
    \caption{\textbf{User Interface for Fine-Grained Interpretation-Quality Annotation.} Users view a pairwise comparison, the model's winner judgment, and its interpretation. In this example, the model is SemJudge, and we render HSGs on the web front end. Users can click a node to view its detailed semiosis (e.g., interpretant or object) and annotate quality in the lower-left panel.}
    \label{fig:fine_anno}
\end{figure*}

\paragraph{User Interface}
We present screenshots of the 2AFC interface, VQA interface, and the fine-grained interpretation-quality interface in Figure~\ref{fig:user_2afc}, Figure~\ref{fig:user_vqa}, and Figure~\ref{fig:fine_anno}, respectively. On average, users spend $14.4$ seconds on a 2AFC task, while experts spend $43.0$ seconds on a VQA annotation and $36.5$ seconds on a fine-grained interpretation-quality annotation; we did not record the time for non-expert VQA annotation.

\subsection{Interpretation Quality Questionnaire}
Participants rated generated interpretations on a 5-point Likert scale (1 = strongly disagree, 5 = strongly agree) using the following dimensions:
\begin{itemize}
    \item \textbf{Causal Agreement (Expert only).} Do the factors in the generated interpretation identified as \emph{decisive} for 2-AFC judgment align with what you consider the primary reasons for that judgment, avoiding spurious, hallucinated, or irrelevant cues?
    \item \textbf{Depth.} Does the interpretation transcend literal description (e.g., object/attribute presence or style adherence) to provide an in-depth, meaning-level analysis (e.g., symbolism, metaphors, theological tradition)?
    \item \textbf{Edification for Artwork Comprehension.} Does this interpretation aid you in comprehending what the artwork may attempt to express (i.e., the creator's intent), compared with seeing the image \& prompt alone?
    \item \textbf{Evidence Grounding}. Are the key claims in the interpretation well-supported by citing specific image regions or global features, and/or by explicit content in the prompt?
\end{itemize}

\subsection{Annotation Consistency and Quality Control.}

\paragraph{Annotation consistency measure.}
We collect raw 2AFC judgments (choose A vs.\ B) from both experts and non-experts, but use them for different roles in quality control. For each 2AFC item, the majority judgment of the expert panel serves as the reference answer. Non-expert judgments are then used to estimate whether the task is sufficiently interpretable at the inter-subjective level, rather than reflecting only a niche or idiosyncratic reading. This process yields an inter-annotator agreement of $0.58$ (Cohen's $\kappa$) over the retained benchmark instances.

\section{Appendix: Extended Discussion}
\label{app:discussion}

\subsection{Why Strong Zero-Shot MLLMs Outperform Tuned Baselines}
\label{app:discussion_zeroshot}

A natural question raised by our main comparison is why a strong zero-shot MLLM baseline (for example, Qwen-3.5-9B without HSG in Table~\ref{tab:additive_ablation}) can outperform tuned art-interpretation systems such as GalleryGPT~\cite{bin2024gallerygpt} and ArtiMuse~\cite{cao2025artimuse}. We emphasize that this outcome \textbf{should not} be read as evidence that a more up-to-date model alone is sufficient for art understanding. Rather, it reflects a combination of \emph{task mismatch} and \emph{generalization regime}.

\paragraph{Task mismatch with prompt-conditioned GenArt semiosis judgment.}
GalleryGPT and ArtiMuse are designed for artwork understanding and analysis, typically on \textbf{existing} artworks or curated art datasets. Our benchmark poses a different problem: the evaluator must judge which generated image better fulfills the \emph{intended meaning} of a prompt in a human--GenArt interaction. This requires not only interpreting the artifact, but also relating that interpretation back to the prompt intent and comparing two candidate realizations. In other words, the task is closer to prompt-conditioned semiosis judgment than to standalone artwork description or classification. A strong zero-shot MLLM may therefore be better matched to the inference pattern required by our benchmark, even if tuned models remain strong on their original task distributions.

\paragraph{Generalization beyond fixed training ontologies.}
The proposed SemiosisArt benchmark also stresses open-ended generalization across heterogeneous motifs, traditions, and generated artifacts. Tuned systems are often optimized on art-historical distributions, which can improve in-domain performance but reduce flexibility on novel prompt-conditioned tasks. By contrast, a strong zero-shot MLLM can exploit broader pretraining and adapt at inference time to unfamiliar symbolic, indexical, or cross-cultural settings. This is especially relevant in our benchmark, where many tasks concern canonically grounded but newly generated images rather than memorized historical works. 

\paragraph{Why matched-backbone comparisons matter most.}
For this reason, the fairest evidence for SemJudge is not only comparison against published tuned systems, but also comparison against alternative prompting-based (i.e., zero-shot) evaluators under the same MLLM backbones. In our experiments, these matched-setting baselines (ArtCoT~\cite{jiang2025multimodal}, DSG~\cite{cho2024davidsonian}, and VIEScore~\cite{ku2024viescore}) still underperform SemJudge, indicating that the gain cannot be explained by model scale alone, but rather by the framing of the evaluation task. These zero-shot evaluation baselines ask what is in the image and how its elements relate (DSG and VIEScore), or assess the image's overall aesthetic quality (ArtCoT). This differs from the measurement of semiosis quality that SemJudge aims to capture. We therefore interpret SemJudge's stronger performance as arising primarily from its semiotics-grounded formulation of the task, while acknowledging that stronger backbones improve absolute performance.

\subsection{Bias, Underrepresentation, and Manifestation Risks}
\label{app:discussion_bias}

Our method and benchmark inherit limitations from both the general-purpose MLLMs used to instantiate SemJudge and the finite cultural coverage of SemiosisArt. These limitations should be understood as part of the evaluation context rather than as noise external to the problem: because SemJudge reconstructs meaning through language-mediated interpretation, any imbalance in the underlying model priors or benchmark coverage can directly shape what kinds of meanings are more legible, more favored, or more easily grounded.

\paragraph{Bias associated with general MLLMs.}
SemJudge relies on general-purpose MLLMs to construct HSGs, localize evidence, and perform downstream judgment. Such models are trained on internet-scale corpora that are known to be uneven in language, geography, aesthetics, religion, and cultural representation. As a result, they may privilege motifs, symbols, and visual conventions that are common in dominant online distributions while under-recognizing less documented, region-specific, or historically specialized traditions. In practice, this can manifest as more confident or coherent interpretations for familiar Western or globally popular motifs, and weaker interpretation quality for culturally specific iconographies whose conventions are sparsely represented in pretraining data.

\paragraph{Underrepresentation in the benchmark.}
As noted in the main limitations section, SemiosisArt broadens coverage across Christian, East Asian, Hindu, and Islamic traditions, as well as modern artistic motifs, but still covers only a subset of the full cultural and stylistic space of generative art. This boundary is driven not only by benchmark scale, but by the scarcity of qualified experts who can reliably formulate and validate canonically grounded interpretation tasks across many traditions. Because our construction protocol also requires strong quality control and inter-subjective validation, some traditions, media, and modes of expression remain underrepresented, especially those whose meanings are highly local, weakly standardized, or difficult to annotate without specialized knowledge. Benchmark performance should therefore be interpreted as evidence within the traditions currently covered, rather than as uniform competence across all cultural settings.

\paragraph{How these biases may manifest in evaluation.}
These two sources of limitation can interact. If a tradition is underrepresented in the dataset and also weakly represented in the MLLM's prior knowledge, SemJudge may produce shallow HSGs, ground the wrong evidence, or prefer variants that are visually plausible but semiotically incomplete. More subtly, even when the final binary judgment is correct, the rationale may still privilege more iconic or globally legible cues over culturally specific symbolic or contextual ones. We therefore recommend interpreting SemJudge's outputs not as neutral truths, but as structured model-based readings whose reliability depends on both representational coverage and evidence grounding.

\section{Appendix: Theoretical Justification for Iconicity Mismatch}
\label{app:theory}

This appendix provides a short theoretical justification for the interpretive principle in Section~\ref{sec:demystify}.
Our goal is not to claim that Peircean semiotics yields this principle as a closed-form theorem by itself, but rather to show that it is a faithful modeling consequence of commitments already built into Peirce's account and our computational formulation~\cite{peirce1991peirce,peirce1992essential}.

\paragraph{Step 1: Peircean dependence on grounds.}
In Peircean semiotics, a sign stands for its object only through a \emph{ground}: the basis on which an interpreter takes the sign to refer to something~\cite{peirce1991peirce,peirce1992essential}. In our formulation, this is written as $g=E(s)$ together with the stands-for mapping $\sigma(g;\eta)\to\hat{o}$. Consequently, if two interpreters rely on different grounds when reading the same sign, they need not recover the same immediate object $\hat{o}$.

\paragraph{Step 2: Iconicity as a one-dimensional summary of grounds.}
For the principle in the main text, we summarize one important aspect of the ground configuration by the scalar $\alpha(s,\eta)\in[0,1]$, which measures how strongly interpreter $\eta$ treats sign $s$ as iconic rather than symbolic/indexical~\cite{peirce1992essential}. This abstraction is appropriate for the present argument because conventional GenArt metrics predominantly reward resemblance-based grounds, whereas many artworks intentionally distribute meaning across symbolic and indexical grounds.

\paragraph{Step 3: From ground mismatch to object mismatch.}
Let
\[
m(s^{(n)}) := \left|\alpha(s^{(n)},\eta^{(n-1)}) - \alpha(s^{(n)},\eta^{(n)})\right|
\]
denote the mismatch between the creator's intended iconicity and the subsequent interpreter's recovered iconicity.
When $m=0$, the creator and interpreter agree on the relative role of iconic grounds, so the interpreter is best positioned to recover an immediate object close to the intended one.
As $m$ increases, the interpreter increasingly privileges the wrong kind of ground. By Peirce's triadic account of sign--object--interpretant mediation~\cite{peirce1991peirce,peirce1992essential}, this shifts the stands-for relation $\sigma(g;\eta)$ and thus increases the risk that the recovered immediate object $\hat{o}^{(n)}$ diverges from the intended object trajectory.

\paragraph{Step 4: Linking to semiosis quality.}
Our paper defines semiosis quality through object-level agreement:
\[
Q_{\mathcal{C}^{(N)}} := -\Delta_o(o^{(1)},o^{(N)}),
\qquad
\hat{Q}_{\mathcal{C}^{(N)}}^\eta := -\Delta_o(\hat{o}^{(1)},\hat{o}^{(N)}).
\]
Since increased iconicity mismatch raises the risk of object divergence (Step 3), and semiosis quality is defined as the negative of that divergence, it follows directly that greater mismatch should, all else equal, degrade semiosis quality. This is the directional principle stated in the main text.

\paragraph{Interpretation.}
The principle should thus be read as a modeling consequence of Peircean semiosis under our computational formalization~\cite{peirce1991peirce,peirce1992essential}: if creator and interpreter disagree about whether a sign should be read iconically versus symbolically/indexically, the recovered object is less likely to match the intended object, and semiosis quality should correspondingly decrease.

\section{Implementation Details}
\label{appendix:system-prompt}
\newcounter{appalgorithm}
\renewcommand{\theappalgorithm}{\thesection.\arabic{appalgorithm}}

\subsection{The SemJudge Algorithm}

We present an algorithmic formulation of SemJudge in Algorithm~\thesection.1.

\begin{figure}[h]
\refstepcounter{appalgorithm}
\label{alg:semjudge_core}
\centering
\begin{minipage}{0.98\columnwidth}
\textbf{Algorithm~\theappalgorithm.} SemJudge: Object-Space Semiosis Quality Measure
\vspace{2pt}
\begin{algorithmic}[1]
\REQUIRE Prompt $s^{(1)}$, Candidate Artifacts $s^{(2)}_a, s^{(2)}_b$
\REQUIRE VLM $\mathcal{M}$ (Evaluator)
\ENSURE Reconstructed Semioses $\mathcal{C}^{(2)}_a, \mathcal{C}^{(2)}_b$, Evidence $\mathcal{L}$, Judgment $\hat{y}$

\STATE \COMMENT{\textbf{Stage 1: Reconstruct Prompt Semiosis (Input Space)}}
\STATE Let $p_{\text{in}}$ be the instruction to analyze $\operatorname{HSG}(s^{(1)})$
\STATE $r_1 \leftarrow \mathcal{M}(p_{\text{in}}, s^{(1)})$
\STATE Extract prompt-level semiotic nodes $\mathcal{V}_1$ from $r_1$
\STATE Initialize context $H \leftarrow [(p_{\text{in}}, s^{(1)}), r_1]$

\STATE \COMMENT{\textbf{Stage 2: Reconstruct Artifact Semiosis (Object Space)}}
\STATE Let $p_{\text{out}}$ be the instruction to analyze $\operatorname{HSG}(s^{(2)}_a)$ and $\operatorname{HSG}(s^{(2)}_b)$
\STATE $r_2 \leftarrow \mathcal{M}(p_{\text{out}}, s^{(2)}_a, s^{(2)}_b \mid H)$
\STATE Extract artifact-level semiotic nodes $\mathcal{V}_{2a}, \mathcal{V}_{2b}$ from $r_2$

\STATE \COMMENT{\textbf{Formulate Cascaded Chains}}
\STATE Construct $\mathcal{C}^{(2)}_a \leftarrow [\mathcal{V}_1 \rightarrow \mathcal{V}_{2a}]$
\STATE Construct $\mathcal{C}^{(2)}_b \leftarrow [\mathcal{V}_1 \rightarrow \mathcal{V}_{2b}]$
\STATE Let $\tilde{\mathcal{V}} \leftarrow \mathcal{V}(\mathcal{C}^{(2)}_a) \uplus \mathcal{V}(\mathcal{C}^{(2)}_b)$

\STATE \COMMENT{\textbf{Stage 3: Evidence Grounding and Judgment}}
\STATE Let $p_{\text{judge}}$ be the instruction to compare chains and cite evidence
\STATE $r_3 \leftarrow \mathcal{M}(p_{\text{judge}} \mid H, r_2)$
\STATE Parse judgment $\hat{y} \in \{a, b\}$ from $r_3$
\STATE Extract rationales $\ell_v$ for nodes $v \in \tilde{\mathcal{V}}$ from $r_3$
\STATE Construct evidence set $\mathcal{L} \leftarrow \{(v, \ell_v) \mid v \in \tilde{\mathcal{V}}\}$

\RETURN $(\mathcal{C}^{(2)}_a, \mathcal{C}^{(2)}_b, \mathcal{L}, \hat{y})$
\end{algorithmic}
\end{minipage}
\end{figure}

\subsection{System Prompt}

Below, we provide the complete system prompts used in our experiments, including prompts for HSG construction from the input prompt and generated images, as well as 2AFC judgment.

\begin{table*}[!t]
\caption{Instructions used for VQA question generation and quality control.}
\label{tab:qa_inst}
\centering
\scriptsize
\setlength{\tabcolsep}{5pt}
\renewcommand{\arraystretch}{1.12}
\begin{tabular}{p{0.22\textwidth} p{0.72\textwidth}}
\toprule
\textbf{Instruction Type} & \textbf{Instruction Summary} \\
\midrule
\multicolumn{2}{l}{\textbf{Question generation instructions}}\\
\midrule
Global generation instruction & Generate exactly one image-grounded multiple-choice QA item for iconographic interpretation evaluation. The final question must be answerable from the visible image and bbox references alone, without access to the original prompt or hidden metadata. The stem and choices must avoid directly describing image contents; if bounding boxes are used, refer only to IDs such as \texttt{bbox\_1}. Questions should be difficult for viewers without relevant iconographic or art-historical knowledge. All distractors must be intra-tradition plausible. Return valid JSON with keys \texttt{question}, \texttt{choices}, \texttt{answer}, and \texttt{rationale}; choices must be exactly \texttt{A}, \texttt{B}, \texttt{C}, and \texttt{D}.\\
\midrule
Pair-comparison generation rule & For paired images, the stem must ask why the known winner is better than the other panel, rather than asking which image wins. The two panels should be referenced only as image a and image b, with only a minimal high-level topical hint. The correct answer should identify the strongest visible comparative reason, and distractors should be hard near-miss comparative alternatives in the same semantic family.\\
\midrule
Negative-sample generation rule & For negative samples, the stem must paraphrase the intended user prompt or requested scene and ask what is wrong with the image relative to that request. The correct answer should diagnose the primary mismatch, inconsistency, or failure, while distractors should be plausible competing diagnoses rather than generic claims that the image is fine.\\
\midrule
Follow-up question generation & A second question for the same image or pair must be meaningfully different from the first one. It should not reuse the same queried relation, answer target, or near-duplicate wording, and it must continue to follow bbox-only references and the non-literal wording constraints used in the initial authoring prompt.\\
\midrule
JSON repair prompt & If the model response does not satisfy the required schema, the pipeline issues a repair prompt asking for valid JSON only. This step is a formatting-recovery instruction that preserves the original authoring task while enforcing the required output structure.\\
\midrule
Spatial-Symbolic Localization & Ask which labeled bounding box instantiates a symbolic role, theological function, or iconographic attribute. The correct answer must require knowledge of the symbolic tradition, not just visual recognition. When boxes are used, cite only bbox IDs and do not name or describe their contents. Distractors should be iconographically plausible alternatives from related traditions. Provide exactly four choices with one correct answer.\\
\midrule
Canonical Deviation Detection & Ask what is iconographically or compositionally anomalous about one labeled bounding box relative to another box or to the invoked tradition. The deviation should be interpretively significant rather than stylistic. Use only bbox IDs when boxes are referenced, and keep distractors as plausible but incorrect deviations grounded in the same tradition. Provide exactly four choices with one correct answer.\\
\midrule
Inter-bbox Relational Meaning & Ask what iconographic or theological meaning is conveyed by the relationship or juxtaposition between two or more labeled bounding boxes. The meaning must depend on the relation rather than either box alone. Use only bbox IDs when boxes are referenced. Distractors should be plausible relational interpretations from adjacent traditions or contexts. Provide exactly four choices with one correct answer.\\
\midrule
Attribute Substitution Consequence & Ask a counterfactual question about how a figure's identity, theological role, or narrative meaning would change if the attribute in one labeled bounding box were replaced with another conventional attribute from the same tradition. The question should probe why the attribute matters semantically, not just what changes visually. Use only bbox IDs when boxes are referenced. Provide exactly four choices with one correct answer.\\
\midrule
Missing Element Identification & Ask which conventionally expected iconographic or compositional element is absent given the narrative, figure, or theological program invoked by the image. The absence must be interpretively meaningful. Distractors should be absent elements whose omission would be unremarkable or expected, so the correct answer depends on canonical knowledge. Provide exactly four choices with one correct answer.\\
\midrule
Hierarchy and Salience & Ask which labeled bounding box has the greatest hierarchical or theological significance, or how multiple boxes should be ranked by significance according to the tradition. The answer must follow compositional or theological convention rather than mere visual prominence. When boxes are used, refer only to bbox IDs. Distractors should reflect plausible alternative hierarchies from other traditions or schemas. Provide exactly four choices with one correct answer.\\
\midrule
Anachronism Detection & Ask which labeled bounding box contains an element that is historically, stylistically, or iconographically inconsistent with the period, tradition, or program followed by the rest of the image. The inconsistency must require knowledge of period-specific convention. When boxes are used, cite only bbox IDs. Distractors should point to boxes that may seem anomalous to a naive viewer but are actually tradition-consistent. Provide exactly four choices with one correct answer.\\
\midrule
Most Probable Theme & Ask which high-level theme, doctrine, philosophical topic, canonical literature, theological motif, or narrative concern is most probably expressed by the image as a whole. The answer must be deducible from the visible image rather than the hidden prompt. If boxes are referenced, use only bbox IDs. Distractors should be thematically plausible alternatives from related traditions. Provide exactly four choices with one correct answer.\\
\midrule
Most Probable Mode of Thematic Expression & Ask how the image's central theme is primarily expressed, such as through spatial opposition, directional movement, hierarchical arrangement, chromatic contrast, figure disposition, or structural repetition. The question should target mechanism rather than theme identity. If boxes are referenced, use only bbox IDs. Distractors should each be compositionally plausible expressive mechanisms. Provide exactly four choices with one correct answer.\\
\midrule
Winner Justification Comparison & For a paired comparison, ask why the known winner image better conveys meaning than the other image. The stem should explicitly compare image a and image b while giving only a minimal high-level hint about the topic. Do not ask which image wins. The correct answer should identify the strongest comparative reason visible across both panels, while distractors should be difficult near-miss comparative interpretations. Provide exactly four choices with one correct answer.\\
\midrule
\multicolumn{2}{l}{\textbf{Quality control instructions}}\\
\midrule
Verifier difficulty anchor & The verifier judges each item against an undergraduate art/history student target. Questions should require comparative, relational, symbolic, or diagnostic reasoning rather than a single obvious local cue, but they should still remain answerable from the visible image without hidden prompt text or excessively esoteric knowledge.\\
\midrule
Verifier distractor standard & Distractors must be genuine same-family near misses. The verifier rewrites items whose wrong options are too weak, too easy to dismiss, uniquely less specific than the gold answer, or otherwise not truly confusable under careful image-based interpretation.\\
\midrule
Verifier diversity and rewrite rule & The verifier compares each candidate item with earlier accepted questions for the same image and rewrites items that overlap too heavily in queried relation, reasoning type, or wording. It also rewrites items with a wrong gold answer, broken authoring rules, insufficient difficulty, excessive obscurity, or overly text-recoverable stems.\\
\midrule
Automatic quality control & Automatic filtering removes generated questions that can be answered from surface-level visual cues, language priors, or text-only shortcuts, and filters out items that are too obscure, underdetermined, or reliant on highly esoteric knowledge. The retained questions should require genuine iconographic interpretation from the visible image and preserve plausible distractors within the same tradition.\\
\midrule
Expert quality control & Held-out experts review the automatically filtered items using the same interpretation-centered standard. They retain only questions with a correct gold answer, grounded rationale, appropriate difficulty, and wording that supports fine-grained evaluation rather than guessable recognition or overly subjective reading.\\
\bottomrule
\end{tabular}
\end{table*}

\begin{table*}[!t]
\caption{System prompt for HSG construction from the user sign.}
\label{tab:hsg_user_prompt}
\centering
\scriptsize
\setlength{\tabcolsep}{5pt}
\renewcommand{\arraystretch}{1.12}
\begin{tabular}{p{0.22\textwidth} p{0.72\textwidth}}
\toprule
\textbf{Component} & \textbf{Prompt Content} \\
\midrule
System instruction & You are an expert Computational Semiotician acting as an interpreter. Your task is to analyze a sign, namely the user input prompt (text or text+image), and infer the user's intention as a structured \textbf{Hierarchical Semiosis Graph (HSG)} whose nodes represent triadic semiosis: sign, object, and interpretant.\\
\midrule
Theoretical framework & The prompt follows Peircean triadic semiosis. For the root and each child node, identify: (i) the \textbf{sign} as the relevant textual or multimodal feature, (ii) the \textbf{object} as the target reality or conceptual subject, (iii) the \textbf{interpretant} as the target effect or mental conception, and (iv) the \textbf{expected grounds} connecting sign to object: iconic, indexical, or symbolic. During input analysis, expected grounds are inferred guidelines rather than hard constraints.\\
\midrule
Task instructions & 1. Analyze the prompt holistically to identify the global object, dominant interpretant, and expected grounds. 2. Decompose the prompt into 3--5 critical sub-signs using concise descriptions. 3. State how each sub-sign contributes to the root node, such as elaboration, contextualization, or stylization.\\
\midrule
Output format & Return a valid JSON object only. The root node is \texttt{hsg\_root} with \texttt{node\_id}, a \texttt{semiosis} object containing \texttt{sign\_description}, \texttt{inferred\_object}, \texttt{interpretant}, and \texttt{expected\_grounds}, plus a \texttt{children} list of sub-sign nodes and their \texttt{relation\_to\_root}.\\
\midrule
Input placeholder & The sign is: \texttt{[\$PROMPT]}.\\
\bottomrule
\end{tabular}
\end{table*}

\begin{table*}[!t]
\caption{System prompt for HSG construction from generated artifacts.}
\label{tab:hsg_artifact_prompt}
\centering
\scriptsize
\setlength{\tabcolsep}{5pt}
\renewcommand{\arraystretch}{1.12}
\begin{tabular}{p{0.22\textwidth} p{0.72\textwidth}}
\toprule
\textbf{Component} & \textbf{Prompt Content} \\
\midrule
System instruction & You are an expert Computational Semiotician acting as a visual interpreter. Your task is to analyze a pair of generated images produced from the same prompt and decode their semiotic structure as two HSGs.\\
\midrule
Theoretical framework & As in the input-sign analysis, follow Peircean triadic semiosis. For the root node and each visual child node, identify sign, object, interpretant, and grounds.\\
\midrule
Task instructions & 1. Analyze each generated image holistically to identify the overarching object, dominant interpretant, and primary grounds. 2. Decompose each image into 3--5 visual sub-signs and analyze their triadic semiosis. 3. When a sub-sign is localizable, provide up to three bounding boxes in image coordinates \texttt{[x\_min, y\_min, x\_max, y\_max]} relative to the full image. 4. State how each sub-sign contributes to the global meaning-making, such as contextualization, contrast, or thematic reinforcement.\\
\midrule
Output format & Return two valid JSON objects only, one for image A and one for image B. Each should contain an \texttt{hsg\_root} with root-level semiosis fields, child nodes, optional \texttt{bounding\_box} entries for localizable elements, and \texttt{relation\_to\_root}.\\
\midrule
Input placeholders & The sign is: A: \texttt{[\$IMAGE\_A]} B: \texttt{[\$IMAGE\_B]}.\\
\bottomrule
\end{tabular}
\end{table*}

\begin{table*}[!t]
\caption{System prompt for 2AFC judgment from reconstructed HSGs.}
\label{tab:twoafc_prompt}
\centering
\scriptsize
\setlength{\tabcolsep}{5pt}
\renewcommand{\arraystretch}{1.12}
\begin{tabular}{p{0.22\textwidth} p{0.72\textwidth}}
\toprule
\textbf{Component} & \textbf{Prompt Content} \\
\midrule
Judgment instruction & Given the raw input and the reconstructed HSGs for the user input and the two model outputs, decide which generated image better fulfills the user's intended object in the input semiosis. The HSGs are used as structured evidence for comparison.\\
\midrule
Input placeholders & \texttt{[\$INPUT\_HSG]} \texttt{[\$OUTPUT\_HSG\_A]} \texttt{[\$OUTPUT\_HSG\_B]}.\\
\midrule
Output format & Return a valid JSON object only with fields \texttt{discussion} and \texttt{winner}. The \texttt{discussion} should give the verbatim decision process with reference to the input and output HSGs, and \texttt{winner} must be either \texttt{``A''} or \texttt{``B''}.\\
\bottomrule
\end{tabular}
\end{table*}

\onecolumn

\section{Appendix: Glossary of Mathematical Notations}
\label{app:glossary}

Table \ref{tab:notation_glossary} summarizes the semiotic and computational notations used throughout the paper.

\begin{table}[!ht]
    \centering

    \renewcommand{\arraystretch}{1.3}
    \begin{tabular}{p{0.18\linewidth} p{0.74\linewidth}}
        \toprule
        \textbf{Symbol} & \textbf{Description} \\
        \midrule
        \multicolumn{2}{c}{\textit{Peircean Semiotics (Section 3)}} \\
        \midrule
        $\xi$ & An atomic semiosis, defined as the tuple $(o, s, i)$. \\
        $s \in \mathcal{S}$ & The \textbf{Sign} (or Representamen); the perceptible form (e.g., prompt, image). \\
        $i \in \mathcal{I}$ & The \textbf{Interpretant}; the meaning constructed by an interpreter. \\
        $o \in \mathcal{O}$ & The \textbf{Dynamic Object}; the underlying meaning or intent not directly observable. \\
        $\hat{o} \in \mathcal{O}$ & The \textbf{Immediate Object}; the underlying meaning or intent as interpreted from sign. \\
        $\eta \in \mathcal{H}$ & The \textbf{Interpreter}; an agent (human or model) mapping signs to interpretants. \\
        $g \in \Gamma$ & The \textbf{Ground}; the evidence or basis (e.g., visual features) connecting a sign to an object. \\
        $E(\cdot)$ & Ground extractor function, where $g = E(s)$. \\
$\rho^{n(\cdot)}$ & Reification function that maps an interpretant to a sign in cascaded semiosis. \\
        $\sigma$ & The ``stands-for'' relationship mapping grounds to objects ($\Gamma \to \mathcal{O}$). \\
        $C^{(N)}$ & A cascaded semiosis consisting of a chain of $N$ atomic semioses. \\
        
        \midrule
        \multicolumn{2}{c}{\textit{HGI Evaluation \& SemJudge (Section 4)}} \\
        \midrule
        $Q_{C^{(N)}}$ & Theoretical quality of a semiosis (distance in Object space). \\
        $\hat{Q}^\eta$ & Empirical quality measure as judged by interpreter $\eta$. \\
        $\hat{o}$ & The inferred object, reconstructed via the inverse stands-for relationship $\sigma^{-1}$. \\
        $\Delta_o, \Delta_g$ & Distance metrics in the Object space and Ground (feature) space, respectively. \\
        $HSG(s)$ & \textbf{Hierarchical Semiosis Graph}; a structured representation of meaning units. \\
        $\mathcal{V}, \mathcal{E}$ & The sets of vertices (atomic semioses) and edges (relations) in an HSG. \\
        $\mathcal{L}$ & A collection of evidence groundings (rationales) linked to graph nodes. \\
        $\ell_v$ & Natural language rationale cited by node $v$. \\
        $y$ & Binary preference label output by SemJudge ($y \in \{a, b\}$). \\
        
        \midrule
        \multicolumn{2}{c}{\textit{Analysis Metrics (Section~\ref{sec:experiment})}} \\
        \midrule
        $NI(s)$ & \textbf{Net Iconicity Score}; measures how much a sign relies on resemblance vs. symbolism. \\
        $Icn, Idx, Sym$ & Individual ratings for Iconicity, Indexicality, and Symbolism. \\
        $\Lambda_k$ & Binary indicator of alignment between an evaluator and human judgment for instance $k$. \\
        \bottomrule
    \end{tabular}
    \caption{\textbf{Glossary of Notations}}
    \label{tab:notation_glossary}
\end{table}

\end{document}